\documentclass[sigconf]{aamas}

\usepackage{balance}
\usepackage{booktabs}
\usepackage[ruled,linesnumbered]{algorithm2e}
\usepackage{placeins}
\usepackage{float}      
   \usepackage{booktabs}      % professional tables
   \usepackage{graphicx}      % figures
   \usepackage{subcaption}    % side-by-side subfigures
   \usepackage{siunitx}       % aligned numbers (optional)
   \usepackage{amsmath}
   \usepackage[table]{xcolor} % row highlighting (optional)
    \usepackage{multirow}
\usepackage{xltabular}   % pulls in longtable + tabularx
\usepackage{hyperref} 
\usepackage{tabularx} 
\usepackage{longtable}  
\usepackage{threeparttable}

\setcopyright{ifaamas}
\acmConference[AAMAS '26]{Proc.\@ of the 25th International Conference
on Autonomous Agents and Multiagent Systems (AAMAS 2026)}{May 25 -- 29, 2026}
{Paphos, Cyprus}{C.~Amato, L.~Dennis, V.~Mascardi, J.~Thangarajah (eds.)}
\copyrightyear{2026}
\acmYear{2026}
\acmDOI{}
\acmPrice{}
\acmISBN{}

\title{Cross-Domain Industrial Fault Detection by Causal Mechanism Monitoring}

\author{Dhiraj Neupane}
\affiliation{
  \institution{Deakin University}
  \city{Waurn Ponds, Victoria}
  \country{Australia}}
  \orcid{0000-0001-6548-311X}
\email{s222439738@deakin.edu.au}
\author{Mohamed Reda Bouadjenek}
\affiliation{
  \institution{Deakin University}
  \city{Waurn Ponds, Victoria}
  \country{Australia}}
  \orcid{0000-0003-1807-430X}
\email{reda.bouadjenek@deakin.edu.au}
\author{Richard Dazeley}
\affiliation{
  \institution{Deakin University}
  \city{Waurn Ponds, Victoria}
  \country{Australia}}
  \orcid{0000-0002-6199-9685}
\email{richard.dazeley@deakin.edu.au}
\author{Sunil Aryal}
\affiliation{
  \institution{Deakin University}
  \city{Waurn Ponds, Victoria}
  \country{Australia}}
  \orcid{0000-0002-6639-6824}
\email{sunil.aryal@deakin.edu.au}

\DeclareMathOperator{\sg}{sg}

\begin{abstract}

Unsupervised fault detection in industrial systems is dominated by reconstruction-based methods that monitor individual sensor marginal distributions. This misses coupling faults, where the physical relationship between sensor groups breaks while marginal statistics remain normal. Such faults evade marginal monitoring and persist as latent failures, with direct consequences for system reliability and safety. We propose CMR-Mamba (Causal Mechanism Representation Mamba), which trains per-domain Mamba state-space encoders on healthy data. A causal cross-modal predictor regularises these encoders so that the effect-channel manifold reflects the normal cause-to-effect coupling. Anomalies are scored by $k$-nearest-neighbour distance on this manifold or by the mechanism residual between the observed and the causally predicted effect embedding. We evaluate CMR-Mamba on electromechanical (Paderborn bearings), hydraulic (ZeMA) and cyber-physical (SWaT) coupling-fault domains. Ablations establish two findings. First, $k$-NN manifold scoring, rather than the encoder family, is the dominant source of gain over reconstruction-error scoring, improving baselines by up to $+0.42$ AUROC and exceeding the gain from causal regularisation. Second, aggregate AUROC is saturated by easy faults that any strong method solves, so the methods separate only on the low-separability subset. There CMR-Mamba leads the evaluated baselines on Paderborn artificial defects and on SWaT stealthy attacks, which keep every sensor inside its normal range and which marginal methods detect only at chance. CMR-Mamba therefore offers an interpretable and consistently competitive approach to coupling-fault detection across mechanical, hydraulic and cyber-physical systems. Code and data are available at \url{https://anonymous.4open.science/status/CMR_Mamba_MFD_1177}.

\end{abstract}


\newcommand{\BibTeX}{\rm B\kern-.05em{\sc i\kern-.025em b}\kern-.08em\TeX}

\setcopyright{none}

\renewcommand\footnotetextcopyrightpermission[1]{}

\begin{document}

\pagestyle{fancy}
\fancyhead{}

\maketitle

\section{Introduction} \label{sec_chap6:introduction}
Rotating machinery such as motors, pumps, and turbines forms the operational backbone of modern industrial systems, from manufacturing plants to power generation facilities and transportation infrastructure. Undetected faults in these machines can escalate from minor defects to catastrophic failures, resulting in unplanned downtime, safety hazards, significant economic losses, and, in severe cases, loss of life~\cite{neupane2025comprehensive,lei2020applications}. The development of reliable and timely fault detection methods has consequently become a central concern in both the research community and industrial practice~\cite{lei2020applications,zhao2019deep}. 

Fault detection has progressed from classical signal-processing methods to data-driven deep learning and, where fault labels are scarce, to unsupervised anomaly detection. The dominant unsupervised approach is reconstruction-based in which a model trained on healthy data flags any sample it cannot reconstruct well. These methods share a structural blind spot, because they monitor the marginal distribution of each signal and answer the question ``Does this signal look normal?''. They therefore cannot see faults that change the physical relationship between signals while each individual signal still looks normal.

Consider a bearing developing a micro-crack, one of the classic sub-surface fatigue mechanisms catalogued for rolling bearings~\cite{iso15243}. In the initial stages, the overall vibration amplitude may remain within the normal operating range, and a reconstruction model trained on healthy vibration would report low error. What has changed, however, is the coupling between the motor current and the resulting vibration. In a healthy machine, the vibration response is tightly coupled to the electrical drive current through well-understood physics. Electromagnetic torque generates mechanical motion, which produces vibration through the bearing and housing~\cite{randall2011rolling}. A micro-crack alters this transfer function, introducing new resonance frequencies and changing damping characteristics, without necessarily changing the amplitude envelope of either signal in isolation~\cite{randall2011rolling,smith2015rolling}. The marginal distributions appear unchanged. The causal mechanism has shifted. An analogous failure mode arises in hydraulic systems. A pump developing internal leakage continues to draw similar motor power and to produce pressure at a similar level, yet the transfer function linking motor power input to pressure output changes measurably. The same electrical driving force now produces less hydraulic work, because a portion of the fluid flow bypasses internally. A reconstruction-based anomaly detector trained on pressure signals would report low reconstruction error because the pressure waveform retains its characteristic shape. The fault is invisible in the marginal signal but detectable in the conditional relationship between motor power and pressure output~\cite{helwig2015hydraulic}. Multi-sensor fusion approaches have shown promise in exploiting cross-channel information~\cite{neupane2025multisensor}, yet they still operate within the reconstruction or classification methodology rather than explicitly monitoring inter-channel coupling. Transitioning these passive monitoring tools into autonomous maintenance agents requires explicitly modelling these structural and causal dependencies~\cite{neupane2026fault}.

This observation motivates the central objective of the present work. Rather than asking ``Does this signal look normal?'', we ask ``Does the physical relationship between these signals still follow the healthy pattern?'' We answer it with the Causal Mechanism Representation Mamba (CMR-Mamba) framework, which rests on two design choices that are independent and each beneficial on its own.
% The first concerns how the encoder is trained. 
First, the domain-specific encoders are trained only on healthy data, with reconstruction as the primary loss and an auxiliary cross-channel coupling regulariser that shapes the latent manifold around the physically grounded causal direction between sensors. A cross-channel predictor learns the normal coupling function from a designated cause channel to a designated effect channel, with the direction fixed by known physics~\cite{randall2011rolling, helwig2015hydraulic}. In electromechanical systems, motor current drives vibration. In electro-hydraulic systems, motor power drives pump pressure. The framework is agnostic to the physical domain as long as this causal direction is grounded. The second choice concerns how anomalies are scored. At inference, a sample is scored by $k$-nearest neighbour ($k$-NN) distance to a bank of healthy embeddings in the effect encoder's latent space, rather than by reconstruction error. When a fault alters the causal mechanism, the effect embeddings are displaced from the healthy manifold even when the marginal signal statistics look normal. We argue that a class of mechanism-shift faults cannot be detected by any method operating on the marginal distributions alone~\cite{bouman2025autoencoders}, yet remains geometrically isolable through manifold monitoring of these causally-regularised representations.

Realizing this framework at the scale of industrial data presents a computational challenge. Industrial sensors often operate at high sampling rates. The Paderborn University (PU) bearing dataset, for example, records vibration and current signals at 64\,kHz, producing 64,000 data points per second per channel~\cite{lessmeier2016condition}. Processing such long sequences is essential for capturing the full spectral and temporal structure of fault signatures, yet Transformer-based architectures incur $\mathcal{O}(L^2)$ computational cost that renders direct processing of raw high-frequency sequences impractical~\cite{vaswani2017attention,gu2024mamba}. Even at considerably lower sampling rates, sequence lengths remain substantial. The ZeMA Hydraulic System dataset~\cite{helwig2015hydraulic} used in our cross-domain evaluation records motor power and pump pressure at 100\,Hz over 60-second operational cycles, producing 6,000 data points per cycle. Efficient sequential architectures are therefore valuable across a broad range of industrial sensing rates. The integration of selective state-space architectures with causal relational modelling provides the efficient spatiotemporal perception required for autonomous industrial diagnostics~\cite{neupane2026fault}. To address this, we employ Mamba~\cite{gu2024mamba}, a selective state-space model that achieves $\mathcal{O}(L)$ complexity while maintaining strong long-range modelling capability. This enables our per-channel encoders to operate directly on raw waveforms at the native sampling rate of each dataset without downsampling or time-frequency transformation, preserving diagnostic information that would be lost through conventional preprocessing. Recent works have begun exploring Mamba for fault detection in supervised settings~\cite{yu2025mamba,xia2025bmtm} and for relation-based fault detection~\cite{chen2025shrinkage}, but none have combined state-space encoders with causal mechanism regularisation in the unsupervised setting, nor demonstrated that the geometric properties of the resulting manifold enable superior anomaly detection compared to reconstruction-based scoring.

The principal contributions of this paper are as follows:

\begin{enumerate}
    \item \textbf{A causal-mechanism architecture that decouples encoder training from anomaly scoring.} We train encoders with reconstruction loss as the primary objective plus a causal coupling regulariser, and score anomalies by $k$-NN manifold distance rather than reconstruction error. An ablation shows the two changes are independently beneficial and compound when combined. The encoders are built from Mamba state-space blocks, whose $\mathcal{O}(L)$ complexity lets us process raw waveforms at each dataset's native sampling rate without time-frequency preprocessing, across sequences from $1{,}500$ to $8{,}192$ tokens and with weight sharing across physically identical channels. The backbone is an adopted component and an enabling choice rather than a contribution of this work.

    \item \textbf{Primary validation and ablation on the Paderborn bearing dataset.} CMR-Mamba detects the industrially relevant real-damage bearing faults at 0.944 AUROC under 15-fold leave-bearing-out cross-validation, and is competitive with the strongest baselines on the aggregate (0.880 overall, up to a 0.930 per-fold oracle upper bound). An ablation crossing encoder type (plain reconstruction versus CMR regularisation) against scoring rule (reconstruction error versus $k$-NN manifold distance) shows that $k$-NN scoring is the dominant factor ($+0.129$ AUROC) and CMR regularisation adds a consistent further gain ($+0.013$).

    \item \textbf{Cross-domain transfer to a hydraulic system, and a further cyber-physical test.} We apply CMR-Mamba without core redesign to the ZeMA hydraulic dataset, swapping the electromechanical pair (motor current to vibration) for the electro-hydraulic pair (motor power to pump pressure), repeating the full ablation and analysing domain-specific differences in the optimal regularisation. The framework is further evaluated on the SWaT cyber-physical water-treatment testbed~\cite{goh2016swat} in the results, where the causal mechanism residual is the best {of the evaluated detectors} of stealthy attacks that keep every sensor within its normal range and that marginal methods detect only at chance.
\end{enumerate}

\section{Background and Related Work}
\label{sec:related}

Traditional approaches to machinery fault detection have relied on
signal processing techniques applied to vibration and acoustic
measurements. Methods such as envelope spectrum analysis, empirical
mode decomposition, and time-frequency representations including the
short-time Fourier transform and continuous wavelet transform have
formed the foundation of vibration-based diagnostics for
decades~\cite{randall2011rolling}. While effective
for well-characterised fault signatures under controlled conditions,
these methods depend heavily on expert knowledge to select appropriate
frequency bands and interpretive thresholds, limiting their scalability
across diverse machine types and operating environments~\cite{lei2020applications}.

The past decade has witnessed a significant shift toward data-driven
approaches, driven by advances in deep learning. Convolutional neural
networks operating on time-frequency images~\cite{bai2023application},
recurrent architectures modelling temporal
dynamics~\cite{zhao2017lstm}, and more recently Transformer-based models
capturing long-range dependencies~\cite{ding2023stationary} have
progressively improved diagnostic accuracy. However, the vast
majority of these methods are supervised \cite{neupane2025comprehensive}, requiring labelled
examples of each fault type during training. In practice, labelled fault
data is scarce, expensive to collect, and inherently machine-specific,
i.e., a classifier trained on one motor's inner race faults may not
generalise to a different motor or even to the same motor under different
operating conditions~\cite{neupane2024comparative,zhao2019deep}. This
fundamental limitation has created a critical bottleneck for scalable
industrial deployment.

This scarcity of labelled data has motivated growing interest in
unsupervised anomaly detection, where models are trained
exclusively on healthy operational data and must flag deviations from
learned normality at test time~\cite{ruff2021unifying}. The dominant
paradigm within this category is {reconstruction-based} detection.
An autoencoder or similar generative model is trained to compress and
reconstruct healthy signals. At test time, samples that yield high
reconstruction error are flagged as
anomalous~\cite{ruff2021unifying,malhotra2016lstm}. This principle
underpins a broad family of methods, ranging from LSTM
autoencoders~\cite{malhotra2016lstm} and variational
autoencoders~\cite{park2018multimodal} to adversarial reconstruction
frameworks such as USAD~\cite{audibert2020usad} and
TranAD~\cite{tuli2022tranad}, and the state-of-the-art Anomaly
Transformer~\cite{xu2022anomaly}, which detects anomalies through
discrepancies in learned self-attention association patterns. One-class
classification methods such as Deep SVDD~\cite{ruff2018deep} offer an
alternative by learning a compact representation of normality without
explicit reconstruction, though the underlying principle, characterising
the healthy data distribution, remains fundamentally the same. Recent
efforts have also explored semi-supervised~\cite{neupane2024comparative}
and reinforcement learning-based~\cite{neupanelearning} formulations to
reduce reliance on fault labels, yet the reconstruction paradigm
continues to dominate the unsupervised
setting~\cite{ruff2021unifying,bouman2025autoencoders}.

Despite their widespread adoption, reconstruction-based methods share a fundamental limitation, i.e., they operate on the marginal distribution of sensor signals. Recent work by Bouman and Heskes~\cite{bouman2025autoencoders} formally proved that autoencoders can perfectly reconstruct out-of-distribution data through three distinct failure mechanisms, namely identity shortcuts, correlated low-level features between in-distribution and out-of-distribution data, and small latent-space norms for anomalous samples. These results hold for both linear and nonlinear autoencoder architectures and represent structural limitations inherent to the reconstruction-based framework, not isolated edge cases~\cite{bouman2025autoencoders}. Complementary work on graph-level anomaly detection has documented the ``reconstruction flip'' phenomenon, wherein anomalous structures paradoxically yield lower reconstruction error than normal ones~\cite{kim2024rethinking}. By learning to reconstruct each channel's output, or the joint output across channels, these methods implicitly answer the question ``Does this signal look normal?'' This framing is effective when faults produce visible deviations in signal amplitude, frequency content, or statistical properties. However, there exists an important class of faults, particularly early-stage mechanical defects{~\cite{iso15243}}, that alter the relationship between sensors before affecting any individual sensor's marginal behaviour.

A large body of work applies deep learning to machinery fault detection and diagnosis~\cite{zhang2017new,zhang2020deep,neupane2024advanced,jiang2025health}.

As mentioned earlier, most of this work is supervised and therefore depends on labelled fault examples. The unsupervised and label-efficient setting that we target is comparatively underexplored~\cite{neupane2024comparative,pang2021deep,neupanelearning}. Recent literature has actively explored this field, spanning deep transfer learning for bearing fault diagnosis~\cite{wang2020multiscale,wang2023online}, physics-informed diagnosis under domain shift~\cite{lu2024dpicen}, semi-supervised remaining-useful-life assessment of rotating machinery~\cite{zhuang2022semisupervised}, and change-point detection coupled with prognostics~\cite{shi2021dual}. Building on these foundations, several recent studies map directly onto the specific components of our proposed approach. On the causal side, structural-causal and anti-causal formulations have been developed for domain-generalised and mechanism-oriented fault diagnosis~\cite{li2026causalgraph,guo2024cis2n,zhang2024anticausal}. On the architectural side, selective state-space (Mamba) encoders have increasingly been applied to industrial prognostics and diagnosis~\cite{han2026mamba}. Related work also covers multi-sensor fusion that couples motor-current and vibration signals~\cite{guo2023multisensor}, interpretable cross-machine condition monitoring~\cite{yan2026unified}, weakly-supervised health-indicator derivation for early fault detection~\cite{hu2025diffusion}, and joint fault diagnosis with remaining-useful-life prediction~\cite{qi2025multitask}.

Our method also draws on causal structure learning and on efficient sequence models. Classical causal discovery~\cite{spirtes2000causation,zheng2018dags} and recent surveys of causal methods for time series~\cite{assaad2022survey} show how directed dependencies can be recovered from observational data, neural Granger causality~\cite{tank2022neural} learns such dependencies with deep networks, and causal representation learning frames the problem at the level of mechanisms~\cite{scholkopf2021causal}. On the architectural side, structured state-space models~\cite{gu2022efficiently,gu2024mamba} provide linear-time sequence modelling that makes raw high-frequency monitoring tractable. 
These two lines have developed separately. Causal discovery and causal representation learning are used mainly to recover or explain structure, and state-space models are used mainly to make long sequences affordable. The work that does bring causal ideas to fault detection is largely supervised or aimed at domain generalisation. What has not been done is to use a learned cause-to-effect mechanism as the anomaly score itself in a label-free setting. {This study does this by} letting a physically grounded cause-to-effect predictor regularise a linear-time state-space encoder, and by scoring anomalies through the geometry of the resulting manifold rather than through reconstruction error.

\section{Methodology}
\label{sec_chap6:methodology}

\begin{figure*}[tbp!]
    \centering
    \includegraphics[width=\textwidth]{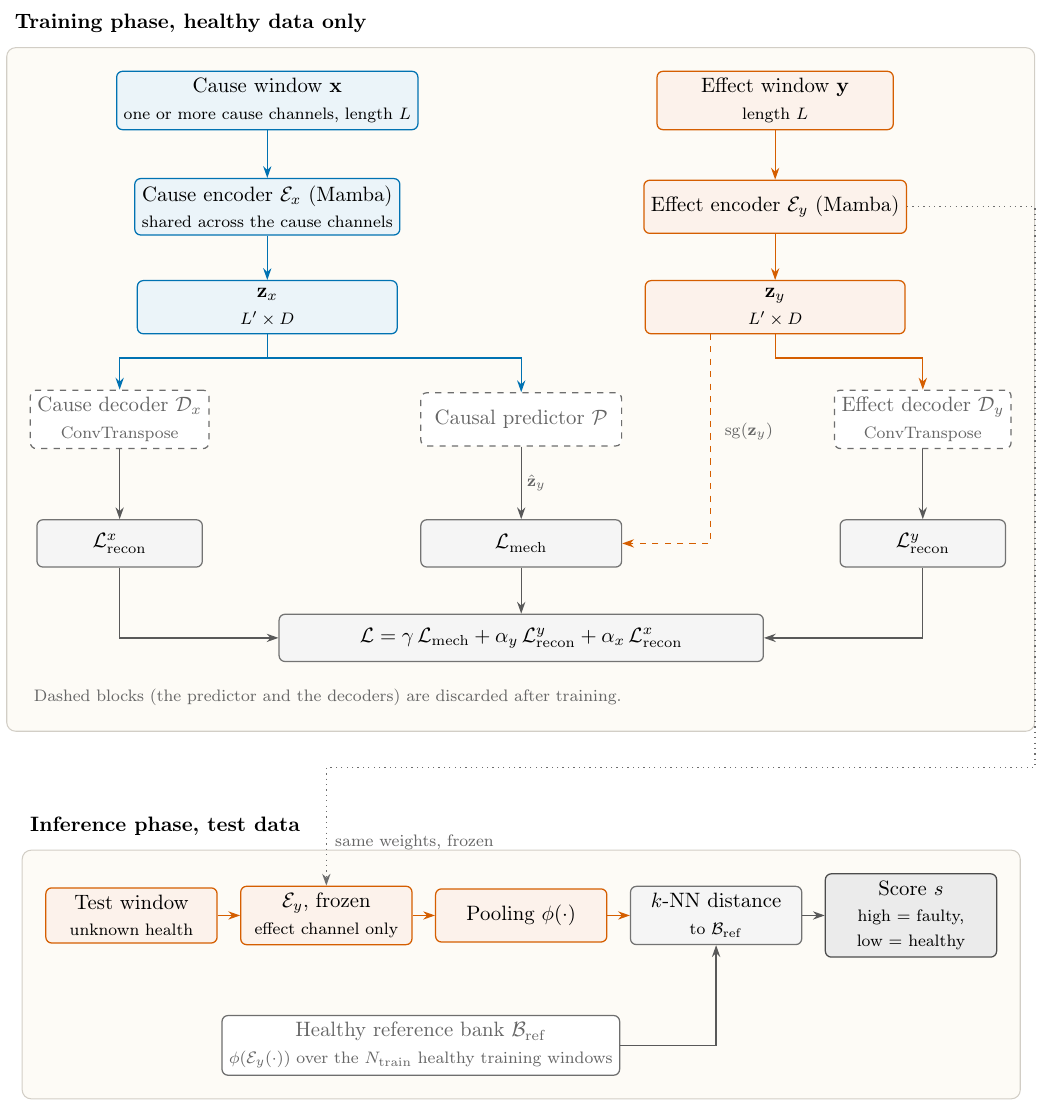}
    
    \caption{{The CMR-Mamba framework, shown generically for a cause channel $\mathbf{x}$ and an effect channel $\mathbf{y}$. In training, on healthy data only, the encoders $\mathcal{E}_{x}$ and $\mathcal{E}_{y}$ map the cause and effect windows to token embeddings $\mathbf{z}_{x}$ and $\mathbf{z}_{y}$. The causal predictor $\mathcal{P}$ predicts the effect embedding from the cause embedding against a stop-gradient target, and the decoders anchor both embeddings to their input waveforms, giving the combined loss $\mathcal{L}$. At inference the predictor and the decoders are discarded, shown dashed, and only the frozen effect encoder is used. A test window is embedded, pooled, and scored by its $k$-nearest-neighbour distance to the healthy reference bank $\mathcal{B}_{\mathrm{ref}}$. Per-domain channel bindings appear in Table~\ref{tab:datasets} and the symbols in Table~\ref{tab:notation}.}}
    \label{fig:cmr_mamba_arch}
\end{figure*}

This section develops the CMR-Mamba framework generically for a cause channel $\mathbf{x}$ and an effect channel $\mathbf{y}$, covering the encoder, the causal predictor and training objective, and the anomaly scoring rule that operate on that pair. The architecture is invariant across domains, with only the input boundary and the channel groupings changing, and each such difference is stated with its physical motivation where it arises. {The PU bearing dataset} is used to develop and ablate the design, ZeMA tests transfer to a physically unrelated domain, and SWaT extends it to a cyber-physical plant. The three datasets, with their preprocessing and splits, are detailed in Section~\ref{sec:experimental_setup}, and Figure~\ref{fig:cmr_mamba_arch} illustrates the general framework implemented in this study.

The framework has three trainable parts, trained together on healthy data only, that shift detection from signal-level appearance to manifold geometry. Domain-specific encoders turn raw sensor channels into sequences of short feature vectors, or tokens, using Mamba blocks so that raw waveforms are processed at the full sampling rate without the quadratic cost of Transformers. A predictor maps the cause token sequence to the effect token sequence, and together with a stop-gradient it shapes the representations around the physical causal coupling. Decoders rebuild the raw waveforms and act as a regulariser that keeps the embeddings informative. The combined loss mixes three goals, rebuilding the cause waveform, rebuilding the effect waveform, and predicting the effect tokens from the cause tokens.

At inference, the predictor and decoders are discarded and only the effect encoder is kept. Every healthy training window is passed through it and pooled to a single fixed-size vector to form a reference bank, and a test window is scored by its distance to the nearest healthy vectors in that bank, so a healthy window scores low and a faulty one high. Section~\ref{subsec_chap6:anomaly_scoring} develops this scoring rule, and Table~\ref{tab:notation} lists the symbols used in the rest of the section.

\begin{table}
\centering
\caption{Notation used in the methodology section.}
\label{tab:notation}
\small
\begin{tabularx}{\columnwidth}{@{}l X@{}}

\toprule
Symbol & Meaning \\
\midrule
$L$ & Raw window length (32,768 for PU, 6,000 for ZeMA, and 256 for SWaT) \\
$L'$ & Token sequence length after the ConvStem ($L/4$) \\
$D$ & Token embedding dimension ($D = 128$) \\
$f_s$ & Sampling rate (64\,kHz for PU, 100\,Hz for ZeMA, and 1\,Hz for SWaT) \\
$\mathbf{x}, \tilde{\mathbf{x}}$ & {Cause-channel window, raw and after z-score normalisation} \\
$\mathbf{y}, \tilde{\mathbf{y}}$ & {Effect-channel window, raw and after z-score normalisation} \\
$\mathbf{z}_{y}$ & Effect-channel token embedding \\
$\mathbf{z}_{x}^{A}, \mathbf{z}_{x}^{B}$ & Cause-channel token embeddings \\
$\hat{\mathbf{z}}_{y}$ & Effect embedding predicted from cause embeddings \\
$\mathcal{E}_{y}, \mathcal{E}_{x}$ & Effect and shared cause encoders \\
$\mathcal{P}$ & Temporal causal predictor \\
$\mathcal{D}_{y}, \mathcal{D}_{x}$ & Effect and cause decoders \\
$\phi(\cdot)$ & Temporal pooling operator \\
$\mathcal{B}_{\mathrm{ref}}$ & Healthy reference bank of pooled embeddings \\
{$s(\mathbf{y})$} & Anomaly score assigned to a test window \\
$\gamma, \alpha_{y}, \alpha_{x}$ & Loss weights for mechanism, effect reconstruction, and cause reconstruction \\
$\sg(\cdot)$ & Stop-gradient operator \\
\bottomrule

\end{tabularx}
\end{table}

\subsection{Encoder Architecture}
\label{subsec_chap6:encoder_architecture}

\begin{figure}[t]
    \centering
    \includegraphics[width=0.49\textwidth]{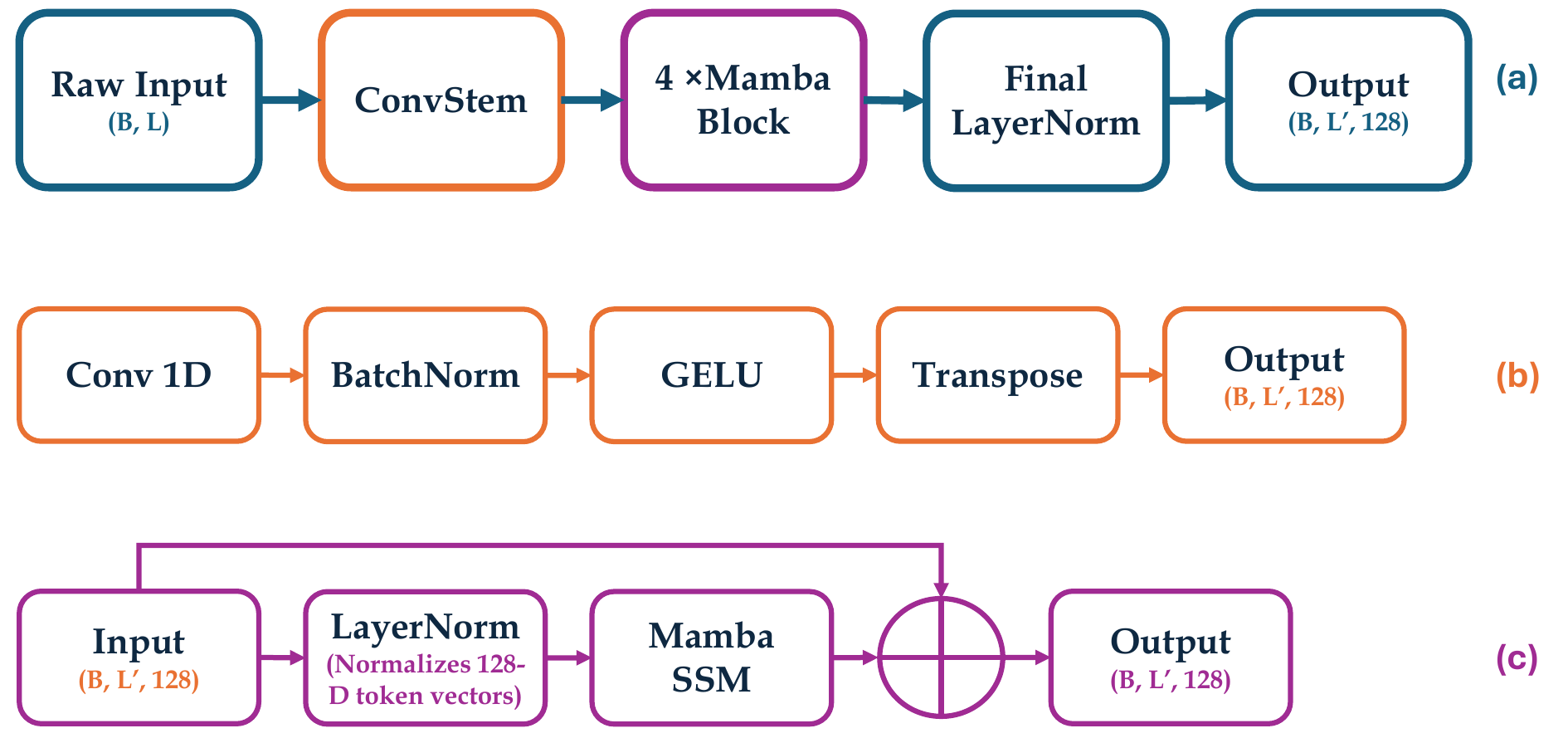}
    \caption{{The per-channel Mamba encoder, drawn generically for an input window of length $L$. (a) The overall pipeline, a ConvStem followed by four stacked Mamba blocks and a final LayerNorm, producing contextual embeddings of shape $(B, L', D)$. (b) The ConvStem, a strided Conv1D with BatchNorm and GELU. (c) A Mamba block, a LayerNorm and Mamba SSM combined through a residual connection. The same encoder is used unchanged on every dataset, with only the input length $L$ and the token count $L' = L/4$ differing across domains (Table~\ref{tab:architecture}).}}
    \label{fig:perChannelMambaEncoder}
\end{figure}

{A single encoder architecture serves all three domains, processing a raw single-channel waveform to return a contextual token sequence (illustrated in Figure~\ref{fig:perChannelMambaEncoder}). Only the input sequence length and the resulting token count vary across datasets while the core architecture remains invariant.}

The encoder is constructed from Mamba selective state-space blocks~\cite{gu2024mamba}, whose linear-time scaling lets it process the raw waveform at each dataset's native sampling rate, which reaches tens of thousands of samples on the highest-rate domain (Table~\ref{tab:datasets}), where a quadratic-complexity architecture would be intractable. Working on the raw waveform matters because the conventional alternatives, downsampling or time-frequency transforms such as spectrograms, discard the fine transient impulse structures that characterise early-stage faults. Mamba is chosen for its selectivity as much as its speed. A time-invariant state-space model would treat transients and background alike, whereas its input-dependent gating stays quiet through the stationary majority of a healthy window and responds sharply at the short transients that carry the fault information.

\paragraph{Convolutional stem:}
{Each encoder begins with a one-dimensional convolution that projects the single-channel input to the $D = 128$ model space using a kernel of seven, a stride of four, and a padding of three. For an input window of length $L$, the output token length is:}
\begin{equation}
    L' = \left\lfloor \frac{L + 2 \times 3 - 7}{4} \right\rfloor + 1 \approx \frac{L}{4},
   \label{eq:convstem}
\end{equation}
{so the stem reduces the raw window to one quarter of its length. Batch normalisation and a Gaussian Error Linear Unit (GELU) activation follow the convolution, and the three stem values follow the physics of the fault transients. A seven-sample kernel is wide enough to span a localized fault impulse yet narrow enough to keep neighbouring transients distinct. The stride of four is the largest downsampling factor that still leaves consecutive kernels overlapping, which gives a kernel-to-stride ratio of $7/4 = 1.75$ and a three-sample overlap so that no input interval falls between tokens. The padding of three is the smallest value that holds the output length at $\lceil L/4 \rceil$. Because the stem indexes samples rather than absolute time, these three values transfer unchanged across domains, and the physical duration a kernel spans in each domain follows from its sampling rate (Section~\ref{sec:experimental_setup}).}

\paragraph{Mamba blocks:}
{The stem output passes through four Mamba layers in a pre-normalisation residual stack, each with an internal state dimension of sixteen, a depthwise convolution width of four, and an expansion factor of two. A final layer normalisation closes the encoder. Each layer updates its representation by}
\begin{equation}
\mathbf{h}_{\ell+1} = \mathbf{h}_{\ell} + \operatorname{Mamba}\bigl(\operatorname{LayerNorm}(\mathbf{h}_{\ell})\bigr),
\label{eq:mamba-residual}
\end{equation}
{where $\mathbf{h}_{\ell}$ is the token sequence at depth $\ell$. The residual path keeps gradients flowing back to the stem, and pre-normalisation holds each block's input variance in check for stable early training. Stacking $N_{\text{enc}} = 4$ blocks gives the encoder as a single composite function}
\begin{equation}
\mathcal{E}(\mathbf{x}) = \operatorname{LayerNorm}\!\Bigl(
   \bigl(\mathrm{Mamba}_{N_{\text{enc}}} \circ \cdots \circ \mathrm{Mamba}_1\bigr)
   \bigl(\operatorname{ConvStem}(\mathbf{x})\bigr)
\Bigr).
\label{eq:encoder}
\end{equation}

{The terminal layer normalisation fixes the cause and effect embeddings to a common scale. Without it, one encoder could drift to a larger variance and dominate the mean-squared mechanism loss of Section~\ref{subsec_chap6:causal_training}, so the predictor would fit a scale mismatch rather than the physical coupling.}

\paragraph{Encoder instances and weight sharing:}
{Encoder allocation follows a single rule. Channels that measure the same physical quantity share one encoder instance, which imposes a common representation as a useful inductive bias, while channels that measure distinct quantities use separate encoders so that one latent space is not forced to span incompatible dynamic ranges. A shared encoder is applied to each of its channels independently and the resulting embeddings are concatenated, which keeps the parameter count controlled when a cause or effect group holds many channels. The concrete channel groups and the sharing pattern for each domain are listed in Section~\ref{sec:experimental_setup}.}

\subsection{Causal Mechanism Prediction and Training Objective}
\label{subsec_chap6:causal_training}

{A single causal predictor maps the cause embedding sequence to a predicted effect embedding, and it is trained against the true effect embedding with a mean squared error under a stop-gradient. Its input and output widths are the only parts that vary between domains, set by the number of cause and effect channels of each (Section~\ref{sec:experimental_setup}). The predictor begins with a fusion layer, a bias-free linear projection followed by layer normalisation, a GELU activation, and dropout of 0.1. Three temporal convolutional blocks with residual connections and causal padding follow, and} {the output projection is initialised to the identity on the two single-coupling domains so that training starts from a stable point (Table~\ref{tab:architecture}).}

\paragraph{Receptive field of the predictor:}
{Three convolutional blocks with kernel five give a receptive field of $1 + 3 \times (5 - 1) = 13$ tokens, which is $13 \times 4 = 52$ raw samples. This window is set to cover the short physical delay between a cause event and its effect response while staying narrow. A wider field was {deliberately avoided}, because excess temporal context would let the predictor memorise the transfer-function fingerprint of individual instances rather than the general mechanism shared across them. The physical delay this field spans in each domain follows from its sampling rate (Section~\ref{sec:experimental_setup}).}

{Causal padding enforces the direction of time. A standard convolution reads tokens on both sides of the target, which would let the predictor use a future cause value to explain a present effect. Restricting the window to the present and the past keeps the learned relationship aligned with the forward direction of physical causation.}

\paragraph{Mechanism and reconstruction losses:}
{The training procedure optimises a composite objective comprising one causal mechanism loss evaluated in the latent embedding space and two reconstruction losses evaluated in the raw waveform space. Defining $\hat{\mathbf{z}}_{y}$ as the predicted effect embedding and $\mathbf{z}_{y}$ as the ground-truth effect embedding, the mechanism loss is formulated as a mean squared error coupled with a stop-gradient operation applied to the target tensor,}
\begin{equation}
\mathcal{L}_{\mathrm{mech}} = \frac{1}{L' D} \bigl\lVert \hat{\mathbf{z}}_{y} - \sg(\mathbf{z}_{y}) \bigr\rVert_2^2.
\label{eq:loss-mech}
\end{equation}
{The operator $\sg(\cdot)$ denotes the stop-gradient constraint, implemented computationally as a single \textit{.detach()} call. This operation preserves the forward pass and the numerical magnitude of $\mathcal{L}_{\mathrm{mech}}$ entirely, acting exclusively on the backward pass to sever the gradient trajectory before it can traverse through the target embedding into the effect encoder. Consequently, the effect encoder optimises solely via its own reconstruction objective, forcing the causal predictor and the cause encoder to actively map toward the stable effect representation rather than allowing the effect representation to drift toward the predictor.} {Without the stop-gradient, the two encoders can co-adapt toward a shared, less discriminative representation, a collapse mode documented for such predictors~\cite{chen2021simsiam}. Its effect is not uniform across the three datasets, and Section~\ref{sec:finding_stopgrad} quantifies where it helps and where it does not.}

{The reconstruction losses are formulated as standard mean squared errors computed over the normalised waveform space. For the effect channel, the loss is defined as}
\begin{equation}
\mathcal{L}_{\mathrm{recon}}^{y} = \frac{1}{L} \bigl\lVert \hat{\mathbf{y}} - \tilde{\mathbf{y}} \bigr\rVert_2^2,
\label{eq:loss-recon-v}
\end{equation}
{where $\hat{\mathbf{y}} = \mathcal{D}_{y}(\mathbf{z}_{y})$ represents the output tensor from the effect decoder.} {The cause reconstruction takes the same form on the cause window,}
\begin{equation}
\mathcal{L}_{\mathrm{recon}}^{x} = \frac{1}{L} \bigl\lVert \hat{\mathbf{x}} - \tilde{\mathbf{x}} \bigr\rVert_2^2,
\label{eq:loss-recon-c}
\end{equation}
{where $\hat{\mathbf{x}} = \mathcal{D}_{x}(\mathbf{z}_{x})$. Paderborn is the one domain with two cause channels, the current phases A and B. They are encoded and decoded by shared-weight instances of $\mathcal{E}_{x}$ and $\mathcal{D}_{x}$, and the cause loss is the average of their two reconstruction errors, $\tfrac{1}{2}(\lVert \hat{\mathbf{x}}^{A} - \tilde{\mathbf{x}}^{A} \rVert_2^2 + \lVert \hat{\mathbf{x}}^{B} - \tilde{\mathbf{x}}^{B} \rVert_2^2)/L$. Both phases carry identical physical information separated only by a phase offset, so an asymmetric weighting would train the shared cause encoder to overfit the dominant phase.}

\paragraph{Decoder:}
{Each decoder inverts the stem as a four-stage upsampler. A feature-mixing convolution is followed by two stride-two transposed convolutions that double the length while contracting the channels from 128 to 64 to 32, and a final projection convolution collapses the result to a single-channel waveform. The first three stages use batch normalisation and a GELU activation, while the output stage is linear because the target is z-scored and spans the real line, so a non-linear output would clip the extreme transients that carry the fault information. Two stride-two stages are used instead of one stride-four stage, because a kernel that is not divisible by the stride produces checkerboard artefacts, whereas two stride-two stages tile the tensor evenly.}

\paragraph{Loss weights and their domain dependence:}
{The final training objective represents a weighted summation of Equations~\eqref{eq:loss-mech}, \eqref{eq:loss-recon-v}, and \eqref{eq:loss-recon-c},}
\begin{equation}
   \mathcal{L}_{\text{total}} = \gamma \, \mathcal{L}_{\text{mech}} + \alpha_{y} \, \mathcal{L}_{\text{recon}}^{y} + \alpha_{x} \, \mathcal{L}_{\text{recon}}^{x},
   \label{eq:loss-total}
\end{equation}
utilising the precise coefficient weights detailed in Table~\ref{tab:training_inference}. Each coefficient was set by sweeping one weight while holding the others fixed on the single evaluation split described in Section~\ref{subsec:pu_ablations}, and the resulting operating point is confirmed under the full 15-fold protocol in Section~\ref{subsec:pu_main}.

{Two of the three weights are fixed across domains. The effect weight $\alpha_{y} = 1.00$ governs the encoder whose representation is used at inference, and lowering it would let that encoder drift toward a representation that suits the predictor but ruins the healthy manifold used for distance scoring. The mechanism weight $\gamma = 0.20$ is the smallest, because the mechanism loss lives in the embedding space and is numerically smaller than the waveform losses, yet even at 0.20 it governs the embedding geometry, as the ablations confirm.}

{The cause weight $\alpha_{x}$ is the only one that varies, and its value follows the physics of the cause channel rather than tuning. Where the cause signal is dominated by the drive that causally determines the effect, as motor current is for bearing vibration, reconstructing it is a useful constraint and $\alpha_{x}$ is non-zero. Where the cause signal carries substantial non-causal overhead, as motor power does through drive-electronics losses, reconstructing it would inject that overhead into the predictor and blur the effect manifold, so $\alpha_{x} = 0$. Section~\ref{subsec:zema_beta} examines this domain dependence in detail.}

{One SWaT-specific choice follows from the training objective. Because only eight of the 25 process sensors carry a learnable actuator-to-sensor coupling (Section~\ref{subsec_chap6:data_preprocessing}), the mechanism residual is measured only on those eight, which keeps it from being diluted by uncoupled channels. This measurement is developed together with the scoring rule in Section~\ref{subsec_chap6:anomaly_scoring}.}

\subsection{Anomaly Scoring by Manifold Distance}
\label{subsec_chap6:anomaly_scoring}

{At inference the decoders are discarded and a test window is scored by its distance to the healthy training manifold rather than by a reconstruction error. This choice of scoring rule matters more for detection performance than the choice of encoder, as the ablations of Section~\ref{sec:finding_scoring} show.}

\paragraph{Pooling:}
{Every training window passes through the trained effect encoder to produce a token sequence of shape $(1, L', D)$. The $k$-NN score needs one fixed-size vector per window, so each sequence is reduced by temporal standard-deviation pooling, taking the standard deviation of every feature across time,}
\begin{equation}
\phi(\mathbf{z}_{y})_d = \sqrt{\frac{1}{L'} \sum_{t=1}^{L'} \bigl(z_{y,t,d} - \bar{z}_{y,d}\bigr)^2}, \quad d = 1, \ldots, D,
\label{eq:pool}
\end{equation}
{where $\bar{z}_{y,d}$ is the per-channel temporal mean. The resulting vector has size $D$ whatever the sequence length $L'$, so one scoring routine serves all three domains even though their token counts differ by two orders of magnitude (Table~\ref{tab:architecture}).}

{Standard deviation is used because of how early-stage faults present physically. A healthy window is close to stationary, so each latent feature stays near a stable baseline, whereas a fault adds brief high-energy impulses at regular intervals~\cite{randall2011rolling} that displace the embedding sharply over a few tokens. Those impulses are too short to move the temporal mean, which the long stretches of ordinary rotation dominate, but they raise the standard deviation, which measures exactly such departures from the baseline. This matches the early-stage regime, where the average amplitude still looks normal while the temporal variance has grown. Mean pooling averages the impulses away, and max pooling responds to the single largest activation, which is as likely to come from a sensor glitch as from a fault. Section~\ref{subsec:pu_embedding} compares these alternatives empirically.}

\paragraph{Reference bank and anomaly score:}
{The healthy reference bank consists of the pooled training embeddings collected immediately following the optimisation phase,}
\begin{equation}
\mathcal{B}_{\mathrm{ref}} = \bigl\{\, \phi\bigl(\mathcal{E}_{y}(\tilde{\mathbf{y}}_i)\bigr) \,:\, i \in \mathcal{I}_{\mathrm{train}} \,\bigr\} \subset \mathbb{R}^{D},
\label{eq:bank}
\end{equation}
{where $\mathcal{I}_{\mathrm{train}}$ indexes the training windows. This bank is constructed once and stored statically without further updates to guarantee computationally efficient and deterministic inference. A test window is subsequently scored by its mean distance to its $k$ nearest healthy neighbours in this reference bank.} {Writing $\mathbf{q} = \phi(\mathcal{E}_{y}(\tilde{\mathbf{y}}))$ for the pooled embedding of the test window and $\mathbf{b}^{(1)},\dots,\mathbf{b}^{(k)}$ for its $k$ nearest neighbours in $\mathcal{B}_{\mathrm{ref}}$, the score is}
\begin{equation}
   s(\mathbf{y}) = \frac{1}{k} \sum_{i=1}^{k} d\!\left( \mathbf{q},\, \mathbf{b}^{(i)} \right).
   \label{eq:knn}
\end{equation}

{Geometrically, the pooled healthy embeddings occupy a compact region of the $D$-dimensional latent space, the healthy manifold. A healthy test window lands on or near it, while a faulty one is displaced along directions that normal operation never visits, and the distance to the $k$ nearest healthy points turns that displacement into a single non-negative score.}

\paragraph{Manifold distance against reconstruction error:}
{Reconstruction error is an unreliable score in principle. Bouman and Heskes~\cite{bouman2025autoencoders} show that an optimised autoencoder can rebuild out-of-distribution inputs with low error, so a faulty window may be reconstructed as faithfully as a healthy one. That failure originates in the decoder, and the manifold score avoids it by working only in the representation space of the encoder, where a decoder that happens to reconstruct a fault cannot affect the distance. The score is also non-parametric. It assumes only that faulty windows fall further from the healthy training support, with no assumption that anomalies reconstruct poorly or that scores follow a particular distribution, and distance to that support is the quantity anomaly detection asks for~\cite{ruff2021unifying}. The ablations support this from both directions. Replacing reconstruction-error scoring with $k$-NN distance on an otherwise unchanged autoencoder accounts for most of the gain, and Deep SVDD, the one baseline whose native objective already measures distance to a compact healthy region, is also the one baseline that $k$-NN scoring does not improve (Section~\ref{sec:finding_scoring}).}

{The same geometry marks the limits of the approach. A fault that degrades the cause-to-effect coupling displaces the embedding along a direction the causal regulariser has trained the encoder to expose, so it is flagged even when its effect on the marginal signal is small. A fault that leaves the coupling intact and merely superimposes an additive component, such as an external vibration source bleeding into the accelerometer, leaves the embedding near the healthy manifold and would likely be missed. The Paderborn faults are of the coupling-degrading kind, which is part of why the framework works well in that domain, and probing the additive case would require a broader fault catalogue.}

\paragraph{The predictor residual as an alternative score:}
{Scoring by the internal error of the predictor is intuitive but works less well on the single-coupling datasets. The predictor is fitted on a handful of healthy training bearings, each with a slightly different mechanical transfer function, so it learns an average that matches no individual bearing exactly. On an unseen healthy bearing its error therefore rises simply because that bearing departs from the average, and this bearing-identity noise swamps the small increase an early-stage fault produces. The manifold score avoids the problem because it never asks the predictor to extrapolate a transfer function to a new bearing. It requires only that training place the healthy embeddings in one contiguous neighbourhood. The ablations confirm the choice, with the predictor-residual score falling well below the $k$-NN manifold score on both mechanical datasets.}

\paragraph{Neighbourhood size and metric:}
{The neighbourhood size and distance metric are set per domain by sweeping $k$ against both cosine and $L_2$ distance, and the sweeps themselves are reported in Section~\ref{subsec:zema_ksweep}. The chosen values track the size of the reference bank and the geometry of the fault. Paderborn uses $k = 1$ with cosine distance and is stable across all fifteen folds. Cosine ignores the overall activation scale, which drifts with operating condition, so at $k = 1$ the score is the smallest angular distance to any single healthy reference vector. ZeMA uses $k = 20$ with $L_2$ distance, because its bank holds only 415 cycles, an order of magnitude fewer than Paderborn, and a single neighbour in so sparse a space gives a volatile estimate. The change of metric follows the fault physics, since bearing damage mainly rotates the embedding while pump leakage displaces it isotropically. SWaT uses $k = 5$ with $L_2$ distance and is insensitive to the choice, with AUROC moving only between 0.846 and 0.863 across $k \in [1, 200]$, which indicates that the attack signal is spread broadly through the embedding space rather than concentrated near one neighbour.}

{SWaT supports a second score that the single-coupling datasets do not. Its attacks break the actuator-to-sensor coupling directly, which makes the mechanism residual, the distance between the observed effect embedding and the one predicted from the concurrent actuator states, a diagnostic signal in its own right. It is computed on the eight strongly coupled sensors and reported alongside the manifold distance, and because it is the stronger signal in that domain, the experimental sections treat it as the primary score there.}

\subsection{Algorithmic Summary}
\label{subsec_chap6:algorithm}

{Algorithm~\ref{alg:cmrmamba} unifies the training sequence, reference bank construction, and inference scoring, providing a consolidated view of the complete operational pipeline.}

\begin{algorithm}[t]
\caption{CMR-Mamba: training and inference.}
\label{alg:cmrmamba}
\SetAlgoLined
\KwIn{Healthy training set $\mathcal{T}$; healthy validation set $\mathcal{V}$; test set $\mathcal{Q}$ of mixed windows; loss weights $\gamma, \alpha_{y}, \alpha_{x}$; neighbourhood size $k$ and distance $d$; maximum epoch count $E_{\max}$; batch size $B$.}
\KwOut{Anomaly scores $\{s(\mathbf{y}_{\star}) : \star \in \mathcal{Q}\}$.}
\textbf{Stage 1: Preprocessing.}\;
Compute per-channel z-score statistics $(\mu_c, \sigma_c)$ from healthy recordings; normalise all windows using Eq.~\eqref{eq:zscore}\;
\textbf{Stage 2: Training.}\;
Initialise the encoders, predictor, and decoders, with the predictor output projection set as described in Section~\ref{subsec_chap6:causal_training}\;
\For{$\text{epoch} = 1, \ldots, E_{\max}$}{
  \ForEach{minibatch $\mathcal{M} \subset \mathcal{T}$ of size $B$}{
    Compute cause and effect token embeddings via Eq.~\eqref{eq:encoder}\;
    Compute predicted effect embedding $\hat{\mathbf{z}}_{y}$ from cause embeddings\;
    Compute the waveform reconstructions $\hat{\mathbf{y}}$ and $\hat{\mathbf{x}}$ for each cause channel\;
    Compute $\mathcal{L}_{\mathrm{total}}$ from Eqs.~\eqref{eq:loss-mech}--\eqref{eq:loss-recon-c} with $\sg(\mathbf{z}_{y})$ as the mechanism target\;
    Backpropagate; clip gradient norm to $1.0$; take an AdamW step\;
  }
  Evaluate $\mathcal{L}_{\mathrm{recon}}^{y}$ on $\mathcal{V}$; checkpoint if it has improved; step the learning-rate scheduler\;
}
Reload the best checkpoint by validation $\mathcal{L}_{\mathrm{recon}}^{y}$\;
\textbf{Stage 3: Reference bank.}\;
\ForEach{training window $i \in \mathcal{I}_{\mathrm{train}}$}{
  Compute $\mathbf{r}_i = \phi\bigl(\mathcal{E}_{y}(\tilde{\mathbf{y}}_i)\bigr)$ using Eq.~\eqref{eq:pool}\;
}
Set $\mathcal{B}_{\mathrm{ref}} = \{\mathbf{r}_i\}$ and store on disk\;
\textbf{Stage 4: Anomaly scoring.}\;
\ForEach{test window $\star \in \mathcal{Q}$}{
  Compute pooled test embedding $\mathbf{q}_\star = \phi\bigl(\mathcal{E}_{y}(\tilde{\mathbf{y}}_{\star})\bigr)$\;
 Find the $k$ nearest neighbours $\mathbf{b}^{(1)},\dots,\mathbf{b}^{(k)}$ of $\mathbf{q}_\star$ in $\mathcal{B}_{\mathrm{ref}}$\;
  Set $s(\mathbf{y}_\star) \leftarrow \tfrac{1}{k} \sum_{i=1}^{k} d(\mathbf{q}_\star, \mathbf{b}^{(i)})$}\;

\end{algorithm}

\subsection{Implementation Details}
\label{subsec_chap6:implementation}

{The framework is implemented in PyTorch, with the official \textit{mamba-ssm} library, version 2.2.2, supplying the selective-scan kernel. Training was executed on a shared cluster with heterogeneous GPUs, so the wall-clock figures in Table~\ref{tab:training_inference} are indicative rather than controlled benchmarks. A Paderborn fold takes about three hours, which puts the full fifteen-fold sweep near 45 GPU hours, a ZeMA seed about twenty minutes, and a SWaT seed about $3.3$ hours. The SWaT cost comes from the width of its multi-channel cause and effect groups rather than from sequence length, since its windows are only 64 tokens. Moreover, the parameter counts follow the same pattern (Table~\ref{tab:architecture}). Paderborn and ZeMA are comparable at $1.67$\,M and $1.59$\,M, whereas SWaT is larger at $5.41$\,M, because its predictor maps a 12-channel cause embedding to a 25-channel effect embedding and a separate decoder is instantiated for each sensor.}

{Model checkpoints are selected on the validation effect-channel reconstruction loss. Because inference relies entirely on the healthy manifold generated by the effect encoder, allowing minor fluctuations in the mechanism or cause-reconstruction losses to dictate model selection would optimize total training metrics {at the expense of downstream anomaly scoring.}}

\paragraph{Reproducibility Requirements:}
{Gradient clipping at a maximum norm of $1.0$ is required, because the selective-scan kernel periodically produces large gradient spikes on windows with sharp transients that otherwise destabilise the early epochs and prevent encoder convergence. Implementation notes, including the \textit{mamba-ssm} submodule import used to avoid a deprecated transformers dependency, are provided in the code repository.}

\begin{table*}[tbp!]
\centering

\caption{{Key training and inference settings. The optimiser configuration is identical across domains, while loss weights, epoch limits, and scoring parameters adapt to the specific dataset. The full specification is in Table~\ref{tab:training_inference}.}}
\label{tab:settings_summary}
\renewcommand{\arraystretch}{1.15}
\small
\begin{tabular}{llll}
\toprule
\textbf{Setting} & \textbf{Paderborn} & \textbf{ZeMA} & \textbf{SWaT} \\
\midrule
$\gamma$, $\alpha_{y}$, $\alpha_{x}$ (loss weights) & $0.20,\,1.00,\,0.75$ & $0.20,\,1.00,\,0.00$ & $0.20,\,1.00,\,0.00$ \\
Optimiser & \multicolumn{3}{l}{AdamW, learning rate $5\times10^{-4}$, weight decay $10^{-5}$, grad clip $1.0$} \\
Maximum epochs & 100 & 200 & 150 \\
{Reported score} & $k$-NN distance & $k$-NN distance & mechanism residual \\
Neighbourhood size $k$, metric & $1$, cosine & $20$, $L_2$ & $5$, $L_2$ \\
Runs reported & 15 folds & 3 seeds & 3 seeds \\
Primary metric & AUROC & AUROC & AUROC, point-wise F1 \\
\bottomrule
\end{tabular}
\end{table*}

\section{Experimental Setup}
\label{sec:experimental_setup}

{This section describes the three datasets that instantiate the framework, the preprocessing and split protocol, and the evaluation protocol used throughout the results.}

\subsection{Datasets, Preprocessing, and Splits}
\label{subsec_chap6:data_preprocessing}
\label{subsec_chap6:splitting}
{The framework is evaluated on three datasets that share a common causal structure but span distinct physical domains. Each records a physically grounded cause channel and an effect channel synchronously, allowing the same cause-to-effect mechanism to be learned without modifying the underlying model. The PU dataset represents an electromechanical system in which motor current drives bearing vibration. The ZeMA hydraulic dataset captures an electro-hydraulic rig in which motor power drives pump pressure. SWaT provides a cyber-physical plant in which actuator commands drive process sensors. Table~\ref{tab:datasets} details the instantiation for each domain.} {Each design decision below, the causal pair, the normalisation rule, and the windowing and splitting protocol, is stated as a general principle and then instantiated for each domain.}

\begin{table*}[!t]
\centering
\caption{{Characteristics of the three coupling-fault domains. The underlying detection framework remains invariant across all datasets, while the causal pairs and temporal window lengths are adapted to the specific physical dynamics of each system.}}
\label{tab:datasets}
\small
\renewcommand{\arraystretch}{1.2}
\begin{tabular}{llll}
\toprule
 & \textbf{Paderborn} & \textbf{ZeMA} & \textbf{SWaT} \\
\midrule
Domain          & Electromechanical & Electro-hydraulic & Cyber-physical \\
Cause channel   & Motor current (A, B) & Motor power (EPS1) & 12 active actuators \\
Effect channel  & Bearing vibration & Pump pressure (PS1) & 25 process sensors \\
Sampling rate   & 64\,kHz & 100\,Hz & 1\,Hz \\
Window          & 32{,}768 samples & 6{,}000 samples & 256 samples \\
                & (0.512\,s, 50\% overlap) & (one 60\,s cycle) & (256\,s, stride 128\,s) \\
Fault           & Bearing damage & Internal pump leakage & Process-control attack \\
Healthy data    & 6 bearings, 6{,}727 windows & 489 stable cycles & 3{,}142 windows \\
Faulty data     & 19 bearings, 21{,}337 windows & 960 cycles & 496 attack windows \\
Split           & Bearing-level, 15 folds & Temporal, 3 seeds & Temporal, 3 seeds \\
\bottomrule
\end{tabular}
\end{table*}

\paragraph{The causal pair:}
{Paderborn~\cite{lessmeier2016condition} records vibration from a piezoelectric accelerometer on the bearing housing together with two motor current phases at 64\,kHz. While most bearing benchmarks record vibration alone, the simultaneous capture of current and vibration at identical resolution in the PU dataset supplies the electrical input and the mechanical response as a matched pair. This synchronous pairing makes the dataset uniquely suited for causal evaluation~\cite{neupane2025multisensor}. The rig is an electromechanical drive with a motor, a torque measurement shaft, a type 6203 deep groove ball bearing under test, and a flywheel load. Each bearing is recorded twenty times for approximately four seconds under four operating conditions spanning two speeds (900 and 1500 revolutions per minute), two load torques (0.7 and 0.1 newton metres), and two radial forces (400 and 1000 newtons)~\cite{neupane2025multisensor}.}

{ZeMA~\cite{helwig2015hydraulic} provides the hydraulic analogue. Two channels sampled at 100\,Hz yield 6{,}000 samples per 60-second cycle. EPS1, the motor power in watts, serves as the cause channel. PS1, the pump output pressure in bar, serves as the effect channel. This causal direction follows from first principles. Motor power drives the pump and the pump converts mechanical energy into hydraulic pressure. Internal leakage breaks this coupling by allowing fluid to bypass internally, causing pressure to fall for the same power input. This mechanism makes leakage the exact hydraulic counterpart to the bearing fault rather than an unrelated failure mode.}

{SWaT~\cite{goh2016swat,mathur2016swat} extends this methodology to cyber-physical systems. The plant records 51 channels at 1\,Hz, comprising 26 discrete actuators and 25 continuous sensors. The causal pair is multi-channel because control actuators collectively drive the process sensors. Pruning 14 constant actuators leaves 12 active components to form the cause group. Linear regression under normal operation separates the sensors that carry a learnable physical mechanism from those governed by unpredictable chemical dynamics. The eight sensors with linear $R^2$ values exceeding 0.55 represent this mechanism, while all 25 provide full system coverage. A SWaT attack forces an actuator or spoofs a sensor to contradict the control commands. This creates a broken control-to-process coupling in exactly the sense the framework targets.}

{Because the encoder and predictor architectures are defined strictly in discrete samples, their effective temporal receptive fields scale naturally with each domain's sampling rate. At the 64\,kHz resolution of the Paderborn dataset, the seven-sample stem kernel spans approximately 0.1\,ms, and the 52-sample predictor field covers 0.8\,ms. These durations tightly encapsulate the 0.08 to 0.31\,ms bearing-fault impulse and the sub-millisecond current-to-vibration delay~\cite{randall2011rolling}. Conversely, at the 100\,Hz resolution of the ZeMA dataset, the identical seven-sample kernel spans 70\,ms, natively accommodating the slower pressure transients characteristic of hydraulic leakage. This sample-domain invariance allows a single architectural configuration to adapt seamlessly across vastly different physical timescales.}

\paragraph{Fault populations:}
{Paderborn contains 32 bearing experiments, comprising 6 healthy bearings (K001 to K006) and 26 faulty bearings, of which 12 carry artificially induced damage and 14 carry real damage from accelerated lifetime tests at 3800 newtons and 2900 revolutions per minute~\cite{lessmeier2016condition}. The artificial defects were produced by electrical discharge machining, electric engraving, and drilling, with each method leaving a geometrically clean and distinct defect~\cite{neupane2025comprehensive}. The real damage arises as fatigue pitting, where subsurface cracks reach the raceway and produce sharp periodic impulses, or as plastic deformation, where the contact geometry is altered smoothly without material removal and without those sharp impulses~\cite{iso15243,randall2011rolling}. Nineteen of the 26 faulty bearings are used for evaluation, comprising 8 artificial and 11 real instances. The 7 excluded bearings are redundant duplicates of damage categories already represented in the dataset. The final selection preserves all five damage processes, both individual races and their combination, severity levels 1 to 3, and six specific severity progressions. These progressions hold the damage mechanism and location fixed while varying only the severity, allowing the anomaly score to be evaluated for monotonic response to physical damage extent. The weighting toward real faults, 11 out of 19, ensures the aggregate metric remains representative of true industrial difficulty. Table~\ref{tab:pu_bearings} lists the resulting subset along with the selection rationale for each bearing.}

{ZeMA records 2{,}205 cycles under four independently varied component conditions. This work targets pump internal leakage, which is encoded across three distinct levels. These levels are healthy, weak, and severe~\cite{helwig2015hydraulic}. A stable flag marks cycles that reached thermal and hydraulic equilibrium prior to recording. Only the 1{,}449 stable cycles are retained. This filtering ensures the model processes steady-state operation rather than unpredictable start-up transients. Following this step, 489 pump-healthy stable cycles form the candidate training pool. The test set comprises 960 pump-faulty stable cycles, divided equally into 480 weak and 480 severe instances. SWaT provides seven days of normal operation followed by four days containing 36 documented attacks. The first six hours of the normal run are discarded to account for plant stabilisation.}

\paragraph{Normalisation:}
{A single normalisation rule governs all three datasets. Each channel is z-score standardised using statistics computed exclusively from healthy data, and these parameters are subsequently applied across every split. {Computing normalisation statistics across the entire dataset would allow high-amplitude fault impulses to artificially inflate the global standard deviation. This inflation would subsequently suppress the relative magnitude of the exact transient anomalies the framework is designed to detect. Restricting the statistical estimation exclusively to the healthy distribution prevents this signal attenuation, aligning with established unsupervised anomaly detection protocols~\cite{ruff2021unifying}.} Formally, for a given channel $c$ with mean $\mu_c$ and standard deviation $\sigma_c$ derived from healthy recordings, every sample is replaced according to the following equation}
\begin{equation}
\tilde{\mathbf{x}}^{c} = \frac{\mathbf{x}^{c} - \mu_c}{\sigma_c + \varepsilon},
\label{eq:zscore}
\end{equation}
{where $\varepsilon = 10^{-8}$ guards against division by zero on a flat channel. On Paderborn the statistics are computed from the six healthy bearings, on ZeMA from the 415 training cycles, and on SWaT from the normal training rows.}

\paragraph{Windowing:}
{The temporal window length is dictated by the physical dynamics of each plant rather than by a fixed convention. Paderborn uses 32{,}768 samples, corresponding to 0.512 seconds at 64\,kHz, with a stride of 16{,}384 to provide a 50 percent overlap. At operating speeds of 15 to 25 revolutions per second, this duration captures 8 to 13 full shaft revolutions. Because every classical bearing fault signature is rotational, a window must span enough revolutions for the repetition pattern to appear multiple times. This allows the encoder to model the pattern as a structural feature rather than a random coincidence~\cite{randall2011rolling,dhirajSN_CNN}. The overlap guarantees that an impulse landing near a boundary is fully covered by at least one window, while simultaneously doubling the healthy training set size at no cost beyond computation. This procedure yields a final Paderborn dataset comprising 6{,}727 healthy windows and 21{,}337 faulty windows. ZeMA requires no windowing. Each 60-second cycle represents an independent operating condition, making one full cycle equivalent to one sample of shape $(2, 6{,}000)$. SWaT employs 256-second windows with a stride of 128 seconds. This length was selected based on the attack-duration distribution, which has a median of 444 seconds. The 256-second window fits entirely inside most attacks while remaining short enough to localise them accurately. A window is labelled an attack window if any timestep within it contains an attack. At 1\,Hz, this window spans only 256 raw samples, indicating that the primary challenge on SWaT is channel count rather than sequence length.}

\paragraph{Splits and leakage:}
{A single rigorous principle governs the data splits. No physical machine instance and no adjacent stretch of time may appear in both the training and testing sets. On Paderborn, the physical instance is the bearing itself. Every window originating from a given bearing is assigned entirely to one split. Two windows from the same recording share nearly all of their physical characteristics and differ only in the specific time region they cover. A naive window-level split would allow a model to achieve high accuracy simply by memorising the unique acoustic fingerprint of an individual bearing without ever generalising to a new machine. The strict bearing-level separation removes that evaluation loophole. Among the six healthy reference bearings, we employ a leave-two-out cross-validation strategy, holding out two bearings for validation and training on the remaining four. There are $\binom{6}{2} = 15$ possible pairs, generating fifteen distinct folds. The bearings are enumerated alphabetically to ensure determinism. All nineteen faulty bearings are reserved exclusively for the test set. The two held-out healthy bearings are also added to the test set to evaluate specificity. This protocol demands two simultaneous levels of generalisation. The model must correctly recognise unseen healthy bearings as healthy, and it must assign systematically higher anomaly scores to the unseen faulty bearings. A method mastering only the first requirement would flag nothing, while a method mastering only the second would trigger constant false alarms. The standard deviation across the fifteen folds explicitly measures how sensitive the final performance is to the specific bearings present in the training set.}

{ZeMA and SWaT each represent a single physical plant, meaning no leave-instance-out split is possible. The separation must therefore be temporal. On ZeMA, the 489 healthy stable cycles are ordered chronologically by their cycle index. The first 85 percent, comprising 415 cycles, form the training set. The remaining 15 percent, comprising 74 cycles, form the validation set. All 960 faulty cycles are placed in the test set alongside the 74 validation cycles to ensure the area under the curve metric has access to both positive and negative examples. This temporal ordering is critical because hydraulic oil temperature drifts slowly over the course of an experiment. A random split would place thermally adjacent cycles into both the training and testing sets, creating subtle data leakage through shared thermal states. On SWaT, the temporal separation is naturally embedded in the data collection, as seven attack-free days precede the four attack days. The normal windows are partitioned chronologically into 3{,}142 training windows and 551 validation windows. The test set comprises 3{,}513 windows, of which 3{,}017 are normal and 496 contain at least one attack timestep. For both of these single-plant datasets, robustness is assessed by training three models from independently drawn random initialisations. The mean and standard deviation of the area under the curve are reported across these three runs. For the ZeMA evaluation, results are reported separately for weak and severe leakage to establish a severity gradient.}

\subsection{Experimental Protocol}
\label{subsec_chap6:evaluation}

\paragraph{Metric:}
{The area under the receiver operating characteristic curve (AUROC) is the primary metric on all three datasets. It is the probability that a randomly chosen faulty window scores higher than a randomly chosen healthy one, so it measures the quality of the ranking without committing to a threshold. That suits this framework, because an operating threshold is a deployment decision governed by the plant-specific costs of false alarms and missed detections, which lie outside an algorithmic comparison. The metric can flatter a detector when the positive class is very rare, but the imbalance here is moderate, with roughly $14\%$ attack windows in the SWaT test set, and at that level the receiver operating characteristic and the precision-recall geometry remain coupled~\cite{davis2006relationship}. Point-wise F1 is additionally reported on SWaT for comparability with the existing literature.}

\paragraph{Estimating variance:}
{Robustness and variance are quantified strictly according to the physical structure of each dataset. Because the Paderborn dataset provides six independent healthy bearings, variance is estimated across the fifteen leave-two-out folds defined in Section~\ref{subsec_chap6:splitting}, with each fold training a model from random initialisation. This strict cross-validation is critical; individual training bearings vary in how comprehensively they span the global healthy manifold. A high mean AUROC coupled with a large fold-to-fold standard deviation indicates algorithmic brittleness rather than generalizable stability, a vulnerability that a pooled mean would mathematically obscure. Conversely, ZeMA and SWaT each record telemetry from a single physical plant, precluding cross-instance validation. Variance for these single-plant domains is therefore estimated across three models trained from independent random initialisation seeds.}

\paragraph{Stratified reporting of fault populations:}
{Aggregate AUROC is never evaluated in isolation, as a pooled metric mathematically obscures the operational boundaries that determine algorithmic viability. Each dataset is therefore explicitly stratified. The Paderborn dataset is partitioned by fault origin into artificial and real damage. A detector that successfully identifies geometrically idealised machined defects but fails on natural fatigue damage has merely overfitted a machining signature rather than learning the generalised physics of bearing failure~\cite{lessmeier2016condition,iso15243}. The ZeMA dataset is stratified by fault severity, separating weak from severe internal leakage to verify that the anomaly score scales monotonically with physical degradation. The SWaT dataset is stratified by attack stealth, classifying attacks as stealthy when every individual sensor remains strictly within its historical normal bounds, and blunt otherwise. This specific split isolates the complex coupling-violating attacks that marginal detectors fundamentally ignore. While tracking distinct physical variables, these three stratification strategies serve an identical diagnostic function, i.e., they separate high-separability faults that any baseline algorithm can identify from the low-separability edge cases that genuinely differentiate competing architectures.}

\section{{Results and Analysis}}
\label{sec_chap6:results}

{This section reports the main cross-dataset results, isolates the scoring rule from the encoder, examines the hard fault subsets, presents the ablations, and closes with the cross-domain analysis.}

\label{sec:pu_experiments}
\label{subsec:pu_setup}

{Paderborn is the primary validation domain, because it is the only one of the three that provides independent machine instances and so supports a leave-instance-out protocol, which makes generalisation to unseen bearings the property under test.} {The data, causal pair, windowing, and splits follow Section~\ref{subsec_chap6:data_preprocessing}, and the architecture and training settings follow Table~\ref{tab:architecture} and Table~\ref{tab:settings_summary}, with the operating point $(\gamma,\alpha_{y},\alpha_{x}) = (0.20,\,1.00,\,0.75)$ selected by the ablations of Section~\ref{subsec_chap6:ablations}. Repeating the fifteen folds three times from independent initialisations gives 45 trained models, meaning 45 instances of CMR-Mamba rather than 45 distinct methods, and figures averaged over them are labelled ``15 folds $\times$ 3 runs''. Aggregate figures are the mean over these 45 models, with the standard deviation taken across the 15 folds.}

\subsection{{Main Cross-Dataset Results}}
\label{subsec_chap6:main_results}

\subsubsection{{Paderborn Results and 15-Fold Cross-Validation}}
\label{subsec:pu_main}

\begin{figure*}[t]
  \centering
  \includegraphics[width=0.99\linewidth]{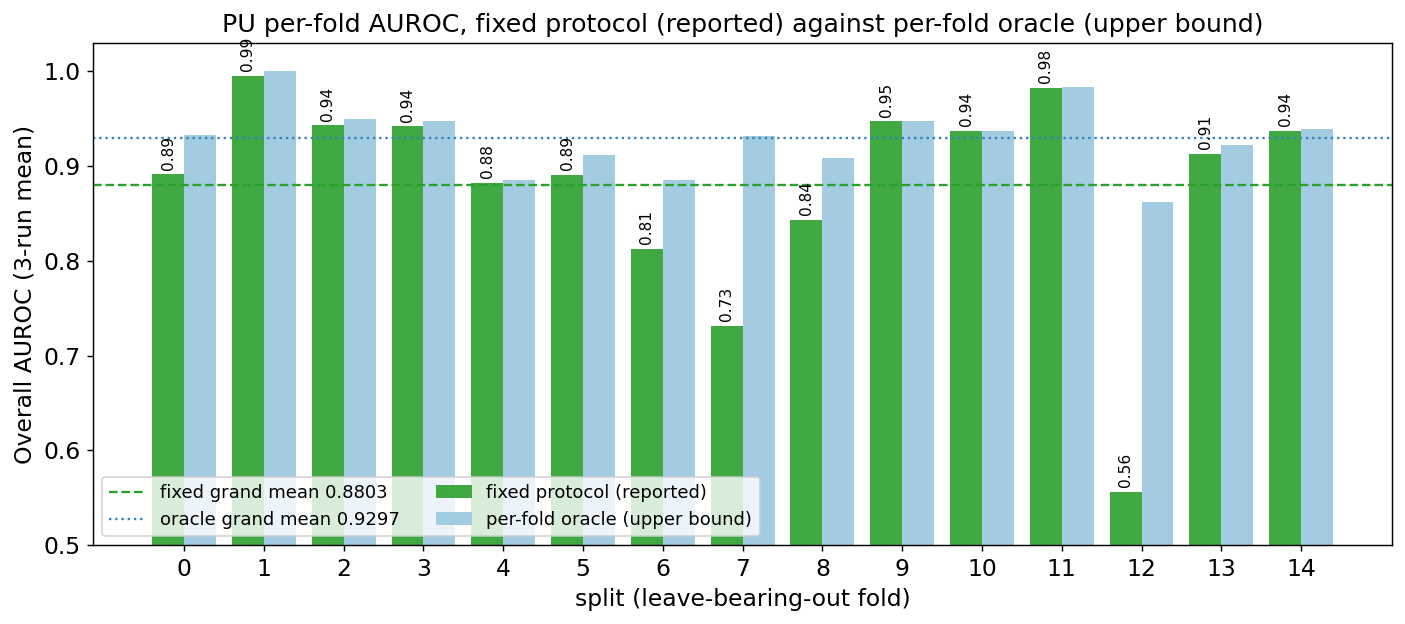}

  \caption{{Per-fold AUROC on PU dataset (3-run mean). The green bars use the fixed evaluation protocol. The light-blue bars select the embedding strategy, neighbourhood size, and distance metric per fold on the evaluation set, which makes them an upper bound. The two series agree on most folds and diverge where per-fold selection can exploit the test set, most sharply on split~12, whose collapse under the fixed protocol is diagnosed below.}}
  \label{fig:pu_main}
\end{figure*}

{Table~\ref{tab:pu_main} reports the per-fold AUROC under the} {fixed loss weights} {$(\alpha_{x},\alpha_{y},\gamma)=(0.75,1.0,0.2)$ averaged over three independent runs. Figure~\ref{fig:pu_main} plots these per-fold means alongside their corresponding run-to-run dispersion metrics.}

{Under the fixed evaluation protocol, a single configuration fixed in advance with std-pooling, $k=1$, and cosine distance applied unchanged to every fold, CMR-Mamba attains a grand-mean overall AUROC of $\mathbf{0.8803 \pm 0.1085}$ across the 15 evaluation folds, decomposing into $0.7919$ on artificial faults and $0.9445$ on real damage. The three runs agree within $0.003$ at the aggregate, so the pipeline is reproducible. Real-damage detection is consistently strong, while the more idealised artificial faults are harder and carry most of the between-fold variance. Evaluated under the same fixed protocol, USAD reaches $0.8850$, the VAE $0.8736$, and the autoencoder $0.8674$ (Table~\ref{tab:pu_fixed_perfold}), so CMR-Mamba is competitive on the aggregate rather than dominant, while remaining the best detector of the artificial-defect subset at $0.7919$ against $0.7538$ for the strongest baseline.}

{Selecting the embedding strategy, the neighbourhood size, and the distance metric per fold on the evaluation set instead raises the grand mean to $0.9297 \pm 0.0351$, with $0.8601$ on artificial faults and $0.9803$ on real damage and run means $0.9313$, $0.9281$, and $0.9297$. This per-fold detail is the series plotted in Figure~\ref{fig:pu_main} and tabulated in Table~\ref{tab:pu_main}. Because it selects hyperparameters on the test folds, it is an upper bound rather than a generalisation estimate and is reported only as a sensitivity bound. The gap between the two protocols is a property of that per-fold selection.}

{Performance approaches near-perfect levels on the most accessible split. Split 1 holds out bearings K001 and K003 and achieves an AUROC of $0.9998$ with a run-to-run standard deviation of only $0.0002$. Conversely performance degrades on a single hard fold. Split 12 holds out bearings K004 and K005 and yields an AUROC of $0.8616$. This specific fold represents the lowest performance across all three independent runs. This systematic difficulty remains intrinsic to the held-out bearing hardware rather than the model architecture. The faulty-to-healthy mean score ratio for split 12 is only $0.97\times$ which indicates that faulty windows score no higher on average than healthy ones. The remaining evaluation folds achieve ratios between $1.4\times$ and $2.8\times$. The healthy embeddings of bearings K004 and K005 intrinsically lie close to the faulty manifold. This overlap represents an irreducible physical property of those two reference components. Reporting the comprehensive per-fold distribution instead of the mean alone provides transparency regarding this hardware limitation and explains the $0.035$ between-fold standard deviation.}

{A direct measurement of the embedding geometry identifies the cause of this fold. Under the fixed protocol, the mean distance from faulty windows to the healthy reference bank, divided by the same distance measured for held-out healthy windows, is $1.045$ on split 12 against a mean of $3.487$ and a minimum of $1.519$ across the other fourteen folds. A ratio of one leaves the score with nothing to separate. The quantity that moves is not the faulty distance, which at $0.120$ is unremarkable, but the held-out healthy distance, which rises to $0.115$ from a typical $0.02$ to $0.05$. Bearings K004 and K005 are the two most atypical of the six on the vibration channel and depart from the population in the same direction on every statistic measured, so they form a mutually covering pair. Whenever one of them remains in the reference bank the other is covered and the fold behaves normally, with held-out healthy distances of $0.027$ to $0.041$ across the six folds that hold out exactly one of them. Split 12 is the only combination of the fifteen that removes both at once, which leaves the bank without coverage of the region those bearings occupy. The collapse is therefore a property of this particular leave-two-out combination rather than of the architecture.}

{Two consequences follow from the diagnosis above. The $0.0048$ gap by which USAD leads CMR-Mamba on the fixed-protocol aggregate is a twentieth of CMR-Mamba's own fold-to-fold standard deviation of $0.1085$, well inside the noise floor of a 15-fold estimate, so it does not support a ranking claim in either direction. The gap is also disproportionately carried by the one fold just diagnosed: over the other fourteen, CMR-Mamba averages $0.9034$ against USAD's $0.8956$, a lead of $0.0078$. This is reported as a decomposition of the aggregate already disclosed above, not as a replacement for it, and split 12 remains in every other figure and table in {this paper}.}

{A Wilcoxon signed-rank test on the fifteen paired folds confirms this assessment. Under the fixed protocol, CMR-Mamba does not differ significantly from USAD ($p=0.64$), the VAE ($p=0.52$), or the autoencoder ($p=0.25$) in the aggregate, while it significantly outperforms Deep SVDD ($p=0.001$). The median paired difference nonetheless favours CMR-Mamba, by $0.015$ over USAD, $0.027$ over the VAE, and $0.016$ over the autoencoder, and CMR-Mamba exceeds USAD on nine of the fifteen folds (Table~\ref{tab:pu_fixed_perfold}).} {USAD's slight edge in the aggregate mean is carried by the single hard fold, split 12, so the fixed-protocol aggregate supports parity rather than a ranking.}

\subsubsection{{ZeMA Results and SWaT Point-wise F1}}
\label{subsec:zema_main}

{The CMR-Mamba framework attains an overall AUROC of $\mathbf{0.9843 \pm 0.0065}$ under the reference configuration utilising no stop-gradient alongside $\alpha_{x}=0$ and $\gamma=0.2$ and $k=20$.} {The no-stop-gradient choice on this dataset is not arbitrary and follows the data-size rule established later in Section~\ref{sec:finding_stopgrad}.} {Table~\ref{tab:zema_main} reports the per-seed performance breakdown alongside the contrasting stop-gradient variant.}

{Two observations follow. First, the framework resolves severe leakage, with the no-stop-gradient severe-fault AUROC averaging $0.9998$ and reaching $1.0000$ on one seed, so nearly all remaining difficulty and variance sits in the weak-leakage faults at $0.9688$ ($k=20$), consistent with a minor leak perturbing the EPS1\,$\to$\,PS1 coupling only slightly. Second, the no-stop-gradient variant is stable across seeds, spanning $0.9752$ to $0.9897$ against the stop-gradient variant's wider $0.8921$ to $0.9858$, a practical advantage for deployment.}

{The SWaT literature frequently evaluates performance using a per-timestep F1 metric operating under the point-adjust protocol. The evaluation maps window scores to individual timesteps to report both the raw best-F1 metric and the point-adjust variant as detailed in Table~\ref{tab:swat_pointf1}{.}}

{This specific evaluation serves two distinct architectural purposes. First the comparison validates the internal evaluation pipeline. The original USAD publication reports a point-adjusted F1 score of approximately $0.79$ on the SWaT dataset. The USAD implementation in this study reaches a raw F1 of $0.774$ alongside a point-adjusted F1 spanning between $0.84$ and $0.90$. These values fall squarely within the published range to confirm the absolute accuracy of the dataset preprocessing and labelling alongside the core evaluation mechanics. Second this analysis exposes the well-documented mathematical pathology inherent to the point-adjust protocol \citep{kim2022towards}. Flagging a single timestep within an attack sequence automatically credits the entire attack segment under this permissive protocol. This generous attribution artificially inflates the performance metric of every evaluated method by margins between $0.06$ and $0.12$ while simultaneously reordering the overarching performance rankings. The $k$-NN point-adjusted F1 score for USAD specifically jumps to $0.899$ under these conditions. {The raw F1 score is therefore the figure reported here.} The point-adjusted variant is provided strictly to ensure empirical comparability with prior literature. This conservative reporting practice is recommended generally for future cyber-physical anomaly detection evaluations.}

\begin{table}[t]
  \centering
  \caption{SWaT point-wise best-F1 (3-seed mean $\pm$ std). Raw F1 is the honest
  metric{, and} point-adjusted (PA) F1 is reported only for comparability.}
  \label{tab:swat_pointf1}
  \begin{tabular}{llcc}
    \toprule
    Model & Score & raw best-F1 & PA-F1 \\
    \midrule
    CMR (NoStopGrad) & $k$-NN  & $\mathbf{0.7769 \pm 0.0017}$ & $0.8470 \pm 0.0039$ \\
    CMR (StopGrad)   & $k$-NN  & $0.7739 \pm 0.0055$ & $0.8882 \pm 0.0039$ \\
    USAD             & native  & $0.7739 \pm 0.0015$ & $0.8359 \pm 0.0019$ \\
    AE               & native  & $0.7733 \pm 0.0043$ & $0.8347 \pm 0.0052$ \\
    VAE              & native  & $0.7730 \pm 0.0013$ & $0.8360 \pm 0.0015$ \\
    USAD             & $k$-NN  & $0.7693 \pm 0.0014$ & $0.8986 \pm 0.0032$ \\
    \bottomrule
  \end{tabular}
\end{table}

\subsubsection{{Consolidated Cross-Dataset Results}}
\label{subsec:synth_scoring}

\begin{figure}[t]
  \centering
  \includegraphics[width=0.99\linewidth]{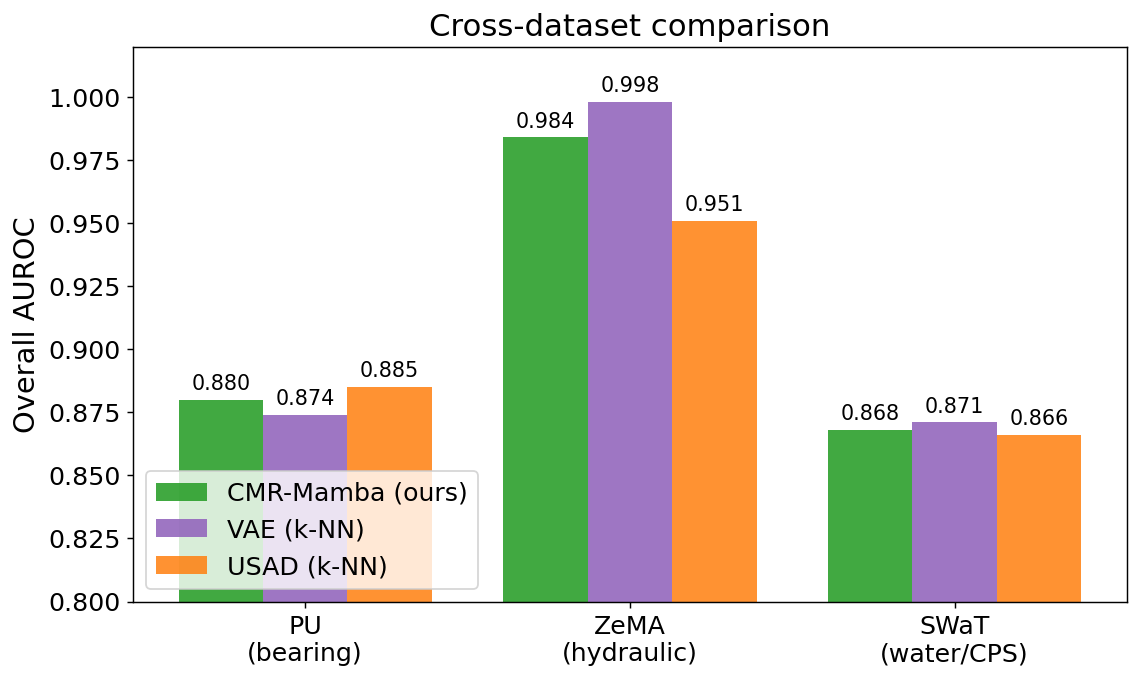}
  \caption{CMR-Mamba against the two strongest baselines (VAE, USAD) across the three domains. {Aggregate scores are close on ZeMA and SWaT, where most faults are easy, and the methods separate on each dataset's hard subset (Table~\ref{tab:hard_subset}).}}
  \label{fig:synth_methods}
\end{figure}

{Table~\ref{tab:synth_master} collects the optimal performance metric of every evaluated method across all three datasets. Figure~\ref{fig:synth_methods} directly compares CMR-Mamba against the two strongest baseline architectures across these domains.}

{Two observations follow. First, no single baseline stays competitive across all three domains. The autoencoder collapses on ZeMA at $0.779$, the Anomaly Transformer is intractable on Paderborn and weakest on SWaT at $0.761$, and Deep SVDD trails on both Paderborn and SWaT at $0.840$. Only the VAE holds up everywhere, and where it exceeds CMR-Mamba the margin sits within overlapping intervals, $0.998$ against $0.984$ on ZeMA and $0.871$ against $0.868$ on SWaT. Second, aggregate AUROC is saturated. It is dominated by high-separability faults that every strong method already solves, with all methods near $\approx 1.0$ on severe ZeMA leakage and a median sensor total-variation distance of $0.61$ on SWaT showing most attacks are blunt. The methods separate on the low-separability hard subset of each dataset, where CMR-Mamba leads (Table~\ref{tab:hard_subset}), so aggregate scoring rewards the easy majority and hides the coupling-aware advantage that disaggregation by difficulty exposes.}

\begin{table*}[t]
  \centering
  \caption{Consolidated overall AUROC. {The PU column applies one configuration
  fixed in advance (std-pooling, $k=1$, cosine) unchanged to all fifteen folds and matched across
  every method. ZeMA and SWaT report each method at its best configuration (3-seed mean).}}
  \label{tab:synth_master}
  \begin{threeparttable}
  % \begin{tabular}{lccc}
  \begin{tabularx}{0.85\textwidth}{@{\extracolsep{\fill}} l c c c}
    \toprule
    Method & PU & ZeMA & SWaT \\
    \midrule
    \textbf{CMR-Mamba (ours)} & {0.880} & 0.984 & 0.868 \\
    VAE                       & {0.874} & 0.998 & 0.871 \\
    USAD                      & {0.885} & 0.951 & 0.866 \\
    AE                        & {0.867} & 0.779 & 0.864 \\
    Deep SVDD                 & {0.663} & 0.846 & 0.840 \\
    Anomaly Transformer       & OOM   & 0.862 & 0.761 \\
    \midrule
    Best CMR score & $k$-NN & $k$-NN & mech.\ resid. \\
    \bottomrule
  \end{tabularx}
  \begin{tablenotes}
    \footnotesize
    \item[] {On PU the four reconstruction methods lie within $0.018$ of one another, and CMR-Mamba's per-fold oracle upper bound there is $0.930$ (Section~\ref{subsec:pu_main}).}
    \item[] {``OOM'' marks the Anomaly Transformer, which is intractable at the PU window length.}
    \item[] {Reconstruction methods are scored by $k$-NN, which improves substantially on their native scores in every domain (Figure~\ref{fig:synth_scoring}), with the per-dataset breakdowns in the appendix.}
    \item[] {Deep SVDD is a centre-distance method, so its native score of $0.892$ on the PU evaluation split is its intended metric. The $0.663$ shown here is its matched $k$-NN value, used to keep the PU column comparable across methods.}
    \item[] {CMR-Mamba uses the no-stop-gradient variant on ZeMA and the stop-gradient variant on Paderborn and SWaT, following the data-size rule of Section~\ref{sec:finding_stopgrad}.}
  \end{tablenotes}
  \end{threeparttable}
\end{table*}

\subsection{{Baseline Comparison and the Role of the Scoring Rule}}
\label{sec:finding_scoring}

{Every evaluation framework is scored twice across all datasets by utilising its own native anomaly metric alongside a $k$-nearest-neighbour distance computed directly on the corresponding embeddings under matched data splits and random seeds. This pairing isolates the structural influence of the scoring rule from the representation capability of the underlying encoder. The Paderborn dataset evaluations are presented first because they exhibit the largest performance delta between the two scoring approaches. The evaluations on the ZeMA and SWaT datasets follow.} {This identical comparative test is then applied directly to the CMR-Mamba encoder itself, and the consolidated comparison across all methods and domains appears in Section~\ref{subsec_chap6:main_results}.} { This empirical effect remains highly consistent across all three domains and constitutes the largest single performance variance measured in this study.}

\subsubsection{{Baseline Comparison on the Paderborn Dataset}}
\label{subsec:base_pu}

{The proposed CMR-Mamba framework is compared against four established one-class baseline models consisting of a convolutional autoencoder, a variational autoencoder, the USAD architecture, and Deep SVDD alongside the Anomaly Transformer. Every baseline reports both its native anomaly score and a separate $k$-NN score calculated using the identical encoder embeddings to isolate the contribution of the scoring strategy from the underlying representation. The consolidated comparison across all methods and datasets is Table~\ref{tab:synth_master}, {and the full per-dataset breakdown with native scores is in the appendix (Table~\ref{tab:pu_baselines}).}}

{Two primary empirical conclusions follow from these results. First the scoring strategy operates as the dominant factor for all reconstruction-based methodologies. Replacing the native reconstruction-error metric with a $k$-NN distance score on the identical embeddings increases the USAD performance from $0.537$ to $0.899$ to achieve a net improvement of $+0.362$. This change similarly raises the variational autoencoder from $0.662$ to $0.906$ for an improvement of $+0.244$ and elevates the standard autoencoder from $0.849$ to $0.907$ to gain $+0.058$. Deep SVDD represents the sole exception to this trend because its hypersphere geometry intrinsically favours its native centre-distance metric yielding $0.892$ compared to a suboptimal $0.646$ under the $k$-NN protocol. Second the three reconstruction encoders are nearly indistinguishable under the unified $k$-NN protocol by spanning a narrow performance range between $0.899$ and $0.907$. The proposed CMR-Mamba framework exceeds the strongest of these baselines by approximately $0.048$ on this specific split. {The single-split baseline figures in this table are not directly comparable with the 15-fold CMR result quoted above, because the latter uses per-fold selection while the baselines use a single fixed configuration. Under the fixed protocol applied identically to all fifteen folds, CMR-Mamba reaches $0.8803$ against $0.8850$ for USAD, $0.8736$ for the VAE, and $0.8674$ for the autoencoder, which places the four methods within $0.018$ of one another.} The Anomaly Transformer cannot execute at the standard Paderborn window length of $32,768$ samples corresponding to $8,192$ tokens because its $\mathcal{O}(N^2)$ self-attention mechanism completely exhausts available GPU memory. This computational bottleneck highlights the practical scalability advantage of the linear-time Mamba encoder architecture at industrial scale.}

\subsubsection{{Baseline Comparison on the ZeMA and SWaT Datasets}}
\label{subsec:base_zema_swat}

{The identical matched comparison framework was executed on the hydraulic leakage task and the cyber-physical testbed. } {Both evaluation domains reproduce the central pattern rather than the exact ordering, with $k$-NN scoring dominating native scoring, the reconstruction encoders clustering tightly, and CMR-Mamba staying competitive with the strongest of them.} {This consistency establishes that the performance differences are not an isolated property of bearing vibration signatures.}

{The four one-class baseline models are evaluated on the ZeMA dataset using identical data splits and random seeds alongside uniform $k$-NN scoring. Each baseline model is assessed at its respective optimal neighbourhood size using a multi-seed protocol to ensure an equal statistical footing {(full detail in the appendix, Table~\ref{tab:zema_baselines}).} A multi-seed variational autoencoder emerges as the single strongest methodology on the ZeMA dataset by achieving an AUROC of $0.9979 \pm 0.0006$ which places it marginally above the CMR-Mamba score of $0.9843$. This performance gap of $0.0136$ remains minor yet consistent. The variational latent space of the variational autoencoder forces the posterior mean to remain smooth and consistent during normal operational cycles. This constraint yields a highly clean manifold structure for subsequent $k$-NN scoring within this single-setting dataset. Conversely the standard autoencoder represents both the weakest baseline and the least stable configuration by yielding an AUROC of $0.7787 \pm 0.1322$. The performance of the autoencoder across different seeds spans a wide range from approximately $0.53$ to $0.83$ whereas the standard deviation of CMR-Mamba is twenty times lower. Deep SVDD achieves its optimal performance at the smallest neighbourhood size of $k=1$ and degrades monotonically as $k$ increases. This trend reflects the specific geometric constraints of its hypersphere optimisation objective. The variational autoencoder is intentionally retained in this comparative analysis. Although it marginally outperforms CMR-Mamba on the ZeMA dataset the broader cross-dataset evaluation demonstrates that no single baseline model remains competitive with CMR-Mamba across all evaluated domains. Omitting the strongest baseline would weaken the overall comparative validity of the framework.}

{The comparative evaluation on the SWaT dataset assesses CMR-Mamba against the autoencoder and the variational autoencoder \citep{kingma2014vae} and the USAD framework \citep{audibert2020usad} and Deep SVDD \citep{ruff2018deep} alongside the Anomaly Transformer \citep{xu2022anomaly}. All models are evaluated under matched conditions utilising the identical windows and splits and random seeds {(full leaderboard in the appendix, Table~\ref{tab:swat_baselines}).} The baseline methods cluster tightly within a narrow performance band between $0.86$ and $0.87$. This close clustering matches expectations from the exploratory data analysis. The median sensor total-variation distance serves as a metric for the separation of two probability distributions across a range from $0$ for identical distributions to $1$ for disjoint distributions. This distance metric computed between the normal and attack marginal distributions of each sensor yields a value of $0.61$. This intermediate value indicates that the majority of SWaT anomalies are blunt attacks that are detectable by any standard model. Aggregate AUROC metrics consequently provide limited separation between the architectures. The variational autoencoder formally achieves the highest value at $0.8706$ followed by CMR-Mamba at $0.8679$. This performance gap of $0.0027$ falls within overlapping confidence intervals and replicates the near-tie conditions observed on the ZeMA dataset. The mechanism residual of the CMR architecture nonetheless outperforms the autoencoder and USAD along with its own $k$-NN score. Only the variational autoencoder edges the proposed framework. The $k$-NN scoring strategy universally dominates native reconstruction scoring across all evaluation sets. This change elevates the reconstruction baselines from approximately $0.77$ under native scoring to approximately $0.86$ under the $k$-NN framework.}

{Deep SVDD yields a lower score of $0.840$ and the Anomaly Transformer trails the performance cluster at $0.761$. The performance of the Anomaly Transformer provides critical insights regarding architectural selection. This attention-based model is computationally intractable on the Paderborn dataset due to its $\mathcal{O}(L^2)$ attention mechanism which exhausts available GPU memory at an $8{,}192$-token window length. The SWaT dataset utilises a short window length of 256 steps which allows the Anomaly Transformer to execute without memory pressure. It nonetheless functions as the weakest method on this dataset. This lower performance demonstrates that even where self-attention remains computationally tractable it is not competitive with the alternative architectures. This empirical result validates the choice of a linear-time state-space backbone based on accuracy considerations rather than memory limits alone. The Anomaly Transformer also represents the single method where the $k$-NN scoring protocol fails to improve upon the native score. Both metrics remain flat across all evaluated $k$ values at approximately $0.76$. The association-discrepancy objective optimisation shapes the resulting attention maps rather than generating a distance-friendly embedding manifold. The $k$-NN manifold trick that systematically elevates the reconstruction baselines is therefore inapplicable to this attention-centric architecture.}
\subsubsection{{Reconstruction Scoring of the CMR Encoder}}
\label{subsec:base_ownencoder}

{This contrast sharpens when the identical evaluation protocol is applied directly to the proposed CMR encoder. Evaluating the CMR encoder via reconstruction error yields an AUROC of only $0.685$. This value falls substantially below both its own $k$-NN score of $0.955$ and the plain autoencoder reconstruction score of $0.849$. This outcome occurs by design rather than representing an architectural deficiency. The causal regulariser intentionally shapes the encoder toward a compact healthy manifold instead of prioritising faithful waveform reconstruction. Reconstruction error consequently becomes a poor anomaly signal. The resulting manifold simultaneously becomes the most discriminative representation among all evaluated encoders. Identical network weights thus yield an AUROC of $0.685$ under decoder-based scoring alongside $0.955$ under decoder-free $k$-NN scoring. The binding operational constraint is therefore the anomaly scoring rule rather than the underlying representation. Even the strongest encoder in this study fails under pure reconstruction scoring. The true detection value of the representation is recovered exclusively through manifold distance evaluation.}

\begin{figure}[t]
  \centering
  \includegraphics[width=0.92\linewidth]{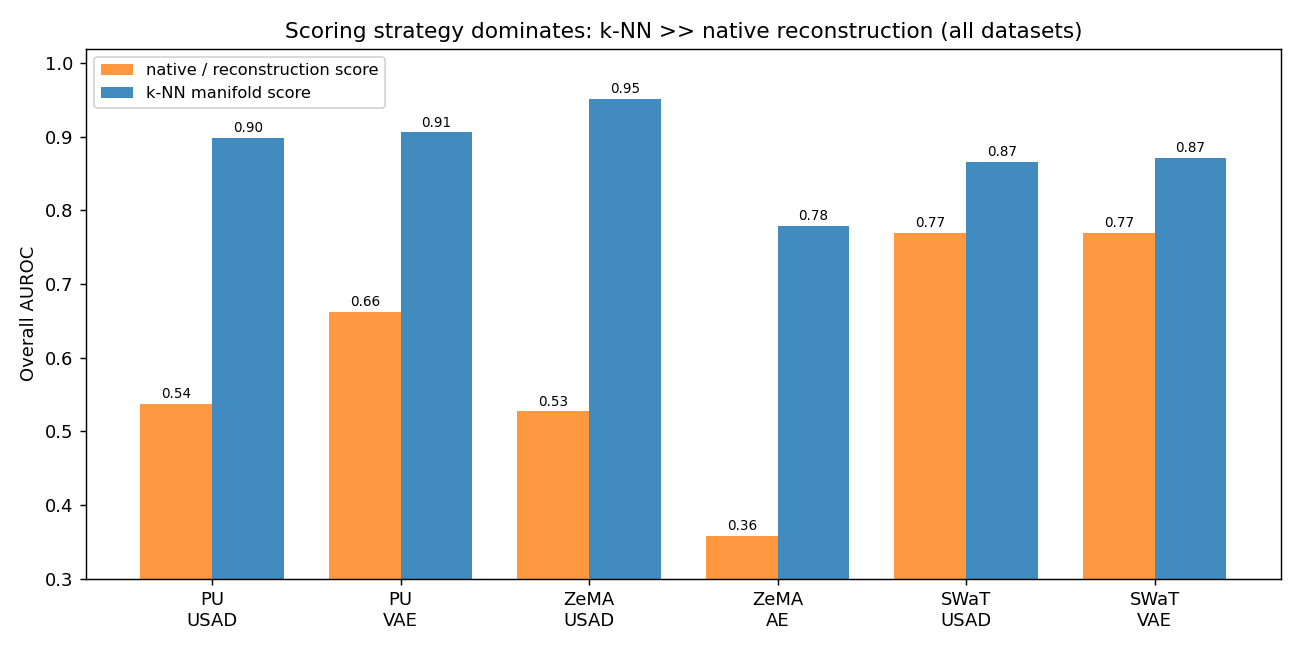}
  \caption{$k$-NN manifold scoring versus native reconstruction scoring across
datasets and methods.}
  \label{fig:synth_scoring}
\end{figure}

{The single largest performance variation observed in this study stems from the choice of anomaly score rather than the choice of encoder architecture. Replacing the native reconstruction or association score of a baseline method with a $k$-nearest-neighbour distance on the identical embeddings improves the AUROC substantially and consistently as illustrated in Figure~\ref{fig:synth_scoring}. The USAD performance on the Paderborn dataset rises from $0.537$ to $0.899$ to yield a $+0.362$ improvement. The variational autoencoder on the same dataset rises from $0.662$ to $0.906$ to achieve a $+0.244$ gain. The standard autoencoder on the ZeMA dataset rises from $0.358$ to $0.779$ representing a $+0.421$ increase. All reconstruction baselines on the SWaT dataset similarly rise from approximately $0.769$ to approximately $0.86$ for a net $+0.09$ gain. The absolute performance gain derived from $k$-NN scoring ranges from $+0.09$ to $+0.42$. This scoring improvement completely dwarfs the architectural gain achieved by adding the causal regulariser to a plain autoencoder which spans only $+0.01$ to $+0.18$.}

{This empirical finding carries two primary consequences. From a practical perspective it demonstrates that industrial practitioners deploying reconstruction-based detectors can recover a massive portion of the achievable performance ceiling simply by transitioning to $k$-NN scoring on their existing embeddings without requiring model retraining. From a conceptual perspective it reframes the fundamental contribution of the CMR-Mamba architecture. The causal mechanism does not function primarily by artificially inflating detection scores. The mechanism operates instead by organising the embedding manifold to ensure that critical operational faults are displaced in geometrically salient directions. This structured representation simultaneously furnishes the mechanism-residual score that no standard reconstruction methodology possesses.}

\subsection{The Mechanism Residual on Low-Separability Faults}
\label{sec:finding_hard}

{The consolidated scores in Section~\ref{sec:finding_scoring} separate the architectures by hundredths of an AUROC point on two of the three datasets, which is an artefact of the saturated aggregate rather than equal capability. All models isolate severe ZeMA leakage at $\approx 1.0$, and the SWaT median sensor total-variation distance of $0.61$ shows most attacks drive at least one sensor out of range, so a method that handles only this easy majority scores almost the same as one that also handles the hard cases.}

{CMR-Mamba targets the opposite class, faults where every channel stays within its normal range while the relationship between channels breaks. These coupling violations form the low-separability hard subset of each dataset, and evaluating on them alone is the discriminating test that shows the mechanism residual is a distinct capability rather than another scoring rule. SWaT gives the clearest such test and Paderborn the second.}

{The aggregate near-tie between CMR-Mamba and the VAE hides a qualitative divergence in which anomalies each isolates. Most SWaT attacks are blunt and drive an affected sensor beyond its healthy range, while a critical minority are stealthy and keep every sensor within its normal historical thresholds so the window looks globally on-distribution. Reconstruction and distribution-based methods flag only off-distribution windows and are structurally blind to the stealthy attacks, whereas the mechanism residual instead asks whether the observed effect sensors stay physically consistent with the concurrent actuator commands.}

{A model-free marginal deviation is calculated for each test window by isolating the maximum absolute time-averaged z-score across the $25$ effect sensors to quantify how far any single channel strays from its historical healthy range. Attack windows that fall below the $95$th percentile of the healthy marginal deviation are formally labelled as stealthy due to their marginal indistinguishability from normal operations. The remaining windows are categorized as blunt. This classification split yields $192$ stealthy attack windows and $304$ blunt attack windows. Every method is subsequently evaluated by calculating the resulting AUROC against healthy windows separately across these two distinct subsets as documented in Table~\ref{tab:swat_stealthy}{.}}

{The disaggregated results justify the causal formulation in a way the aggregate hides. On the stealthy subset the CMR mechanism residual leads at $0.747$, ahead of the VAE at $0.701$ and every other baseline, while the pure-marginal scores collapse below chance to about $0.40$, confirming these coupling anomalies are invisible to marginal detectors. The ordering reverses on the blunt subset, where marginal and reconstruction methods reach $0.97$ to $0.999$ and the mechanism residual stays effective at $0.949$ but trails them. Because blunt windows outnumber stealthy ones $304$ to $192$, the VAE's edge on the blunt majority produces the marginally higher aggregate and the near-tie of Section~\ref{sec:finding_scoring}.}

{Two conclusions follow. First, the mechanism residual is a distinct detection capability, not an alternative scoring trick, isolating the stealthy coupling-violating attacks that a VAE or autoencoder cannot capture, which are exactly what an adversary would design to evade marginal monitoring. Second, the residual and reconstruction manifold scoring are complementary, so a deployed monitor is best served by both, using reconstruction for blunt anomalies and the residual for stealthy ones. The stealthy subset of $192$ windows warrants caution on precise margins, but the qualitative pattern is unambiguous, with the causal residual best on stealthy attacks and marginal methods best on blunt ones.}

{Because the stealthy subset is defined by a single percentile cut of the model-free marginal deviation, its robustness to that cut matters. Re-splitting the attack windows at the 90th, 95th, and 99th healthy percentiles leaves the ordering unchanged. The CMR stop-gradient mechanism residual leads the stealthy subset at every threshold, at $0.747$, $0.747$, and $0.753$, ahead of the VAE at $0.701$, $0.701$, and $0.708$, while the pure-marginal score and the native VAE stay between $0.40$ and $0.44$ throughout. The 90th and 95th cuts give the identical subset because no attack window falls between the two thresholds, and the 99th cut adds eleven borderline windows without moving the result. The stealthy advantage is therefore a property of the attacks rather than of the chosen threshold.}

\begin{table}[t]
  \centering
  \small
  \caption{Stealthy versus blunt attack detection on SWaT (AUROC against healthy,
3-seed mean). {On stealthy attacks the CMR mechanism residual leads, and on
blunt attacks the marginal and reconstruction methods lead.}}
  \label{tab:swat_stealthy}
  \begin{tabular}{lcc}
    \toprule
    Method & Stealthy (192) & Blunt (304) \\
    \midrule
    
    \textbf{CMR mechanism residual (StopGrad)} & \textbf{0.747} & 0.949 \\
    CMR mechanism residual (NoStopGrad) & 0.709 & 0.947 \\
    VAE ($k$-NN) & 0.701 & 0.978 \\
    USAD ($k$-NN) & 0.693 & 0.974 \\
    AE ($k$-NN) & 0.681 & 0.981 \\
    Marginal deviation / native recon. & $\approx 0.40$ & $\approx 0.999$ \\
    \bottomrule
  \end{tabular}
\end{table}

\subsubsection{Artificial Defect Faults on the Paderborn Dataset}
\label{subsec:hard_pu}

{Paderborn shows the same dichotomy, here set by the physics of the damage rather than an attacker's intent. Real damage produces distributed energetic signatures that every reconstruction baseline detects above $0.97$, while artificial defects are geometrically idealised and perturb the current-to-vibration coupling far more subtly, and they carry the aggregate variance. } {Under the fixed 15-fold protocol CMR-Mamba is the strongest of the learned detectors on the artificial faults, at $0.792$ against $0.754$ for the strongest baseline, USAD, winning ten of fifteen folds against USAD and the VAE, eleven against the autoencoder, and fourteen against Deep SVDD. On the single tuning split the ordering is sharper, at $0.900$ against $0.809$ for the autoencoder. {Real damage is detected reliably by every strong method, the $0.97$ figures above being single-split values, and CMR-Mamba reaches $0.944$ across the fifteen folds, so the methods separate on the artificial-defect subset rather than on real damage. That artificial-subset margin is far larger than the hundredths separating the methods on the aggregate, so the architectures are near-parity on easy faults and diverge on the harder ones.}}

{Table~\ref{tab:hard_subset} compares both hard subsets. ZeMA has no equivalent entry, because its weak-leakage faults, though harder than severe leakage, are still resolved near-perfectly by the strongest baselines, so it offers no subset that separates the methods.}

{The conclusion aligns with the causal claim. The mechanism residual does not inflate scores on easy anomalies, it enables detection of a fault class that evades marginal and reconstruction scoring entirely, the coupling-violating anomalies. That capability adds only a few thousandths to the aggregate but becomes decisive on the attacks an adversary engineers and the subtle defects of early bearing failure, contributing an absolute margin of $0.05$ to $0.09$ on these subsets and fixing the performance hierarchy.}

\begin{table}[t]
  \centering
  \caption{Hard-subset comparison. On each dataset's low-separability faults, the ones the coupling paradigm targets, CMR-Mamba is the best of the evaluated learned detectors. The PU values come from the fixed 15-fold protocol with identical $k$-NN scoring for all methods, and the SWaT values are the 3-seed mean from Table~\ref{tab:swat_stealthy}. ``Marginal'' is the model-free per-sensor deviation score.}
  \label{tab:hard_subset}
  \begin{tabular}{lccc}
    \toprule
    {Hard subset} & {CMR-Mamba} & {Best baseline} & {Marginal} \\
    \midrule

    {PU artificial defects} & {\textbf{0.792}} & {0.754 (USAD)} & {---} \\
    
    {SWaT stealthy attacks} & {\textbf{0.747}} & {0.701 (VAE)} & {$\approx 0.40$} \\
    \bottomrule
  \end{tabular}
\end{table}

\subsection{Ablation Studies}
\label{subsec_chap6:ablations}

\subsubsection{Loss-Weight Ablations on Paderborn}
\label{subsec:pu_ablations}

{The three loss weights in Equation~\eqref{eq:loss-total} were set by single-split ablations on the standard evaluation split of about $4{,}486$ healthy training windows and $23{,}578$ test windows, then validated under the full 15-fold protocol. Single-split values carry the per-fold noise quantified in Section~\ref{subsec:pu_main}, a 15-fold standard deviation of about $0.035$, so they are read as trends. The weights were tuned by coordinate descent from a working point, sweeping one weight while the other two were held fixed. The reconstruction weights $\alpha_{x}$ and $\alpha_{y}$ anchor the encoders to their inputs and are set first, and $\gamma$ is swept afterwards at $\alpha_{x}=0.75$, $\alpha_{y}=1.0$. The stop-gradient is structural rather than a tunable weight and is evaluated {separately below}.}

\paragraph{{Reconstruction Weights \texorpdfstring{$\alpha_{x}$ and $\alpha_{y}$}{alpha-curr and alpha-vib}}}
\label{subsec:pu_alpha}

{Each reconstruction weight is swept as a one-dimensional slice averaged over three runs (
% Table~\ref{tab:pu_alpha} and 
Figure~\ref{fig:pu_currvib} in the appendix), with $\alpha_{x}$ varied at $\alpha_{y}=1.0$ and then $\alpha_{y}$ varied at the selected $\alpha_{x}=0.75$. The current weight peaks at $\alpha_{x}=0.75$ with $0.9573 \pm 0.0031$, and $\alpha_{x}=0$ drops it to $0.9393$ because the cause encoder then sees no reconstruction signal, while the range $0.25$ to $1.0$ stays within $0.942$ to $0.957$. The vibration weight collapses to $0.8889$ at $\alpha_{y}=0$, confirming the effect encoder needs a reconstruction anchor, and is otherwise flat, with $\alpha_{y}\in\{0.5,0.75,1.0\}$ within $0.005$ of one another at $0.9396$, $0.9457$, and $0.9466$. We fix $(\alpha_{x},\alpha_{y})=(0.75,1.0)$, matching the vibration channel's role as the monitored effect.}

\paragraph{{Mechanism Loss Weight \texorpdfstring{$\gamma$}{gamma}}}
\label{subsec:pu_gamma}

{The weight $\gamma$ controls how strongly the causal-mechanism term shapes the learned embedding relative to signal reconstruction. The parameter $\gamma$ is evaluated across the set $\{0.0, 0.05, 0.1, 0.2, 0.5, 1.0\}$ while maintaining $\alpha_{x}=0.75$ and $\alpha_{y}=1.0$ as constant values. Table~\ref{tab:pu_gamma}
% and Figure~\ref{fig:pu_gamma} 
reports these results.}

{The response is unimodal and peaks at $0.9405$ at $\gamma=0.2$, which we adopt. Smaller weights under-enforce the mechanism and leave the embedding close to a plain reconstruction, for example $0.9203$ at $\gamma=0.05$, while larger weights over-constrain it against reconstruction fidelity, $0.9229$ at $\gamma=0.5$. The overall AUROC stays within a $0.021$ span across the whole range, and the real-damage score is near-saturated between $0.984$ and $0.995$ throughout, so $\gamma$ acts mainly on the harder artificial-fault subset.} {Both the $\alpha_{x}$ slice and this $\gamma$ slice were run at $\gamma=0.2$, so their shared nominal point $(\alpha_{x},\alpha_{y},\gamma)=(0.75,1.0,0.2)$ is the same configuration, and the small difference between the two reported values ($0.9573$ against $0.9405$) is single-split run-to-run variation between two independent coordinate-descent passes. The operating point is confirmed under the 15-fold protocol in Section~\ref{subsec:pu_main}.}

\subsubsection{{Embedding Strategy and Neighbourhood Size on Paderborn}}
% \subsection{{Embedding and Neighbourhood Choices}}
\label{subsec:pu_embedding}

\begin{figure}[!htbp]
  \centering
  \includegraphics[width=0.99\linewidth]{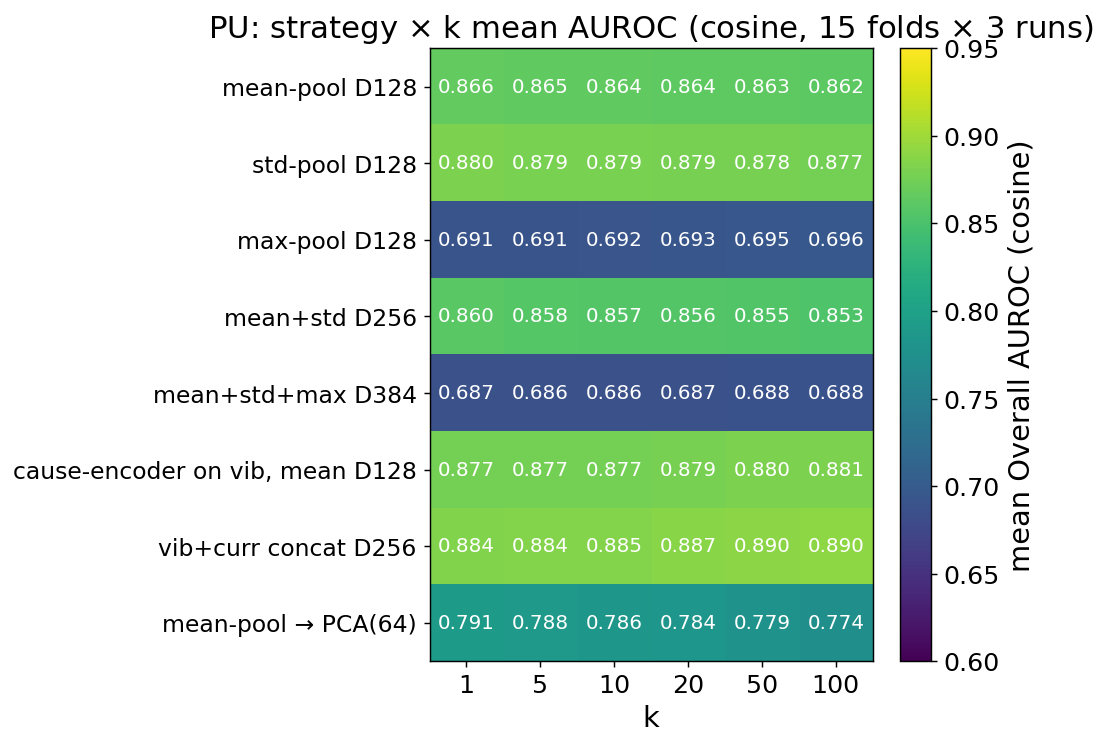}
  \caption{PU strategy $\times\,k$ mean AUROC {(cosine, 15 folds $\times$ 3 runs)}. The \textit{max}-based
{descriptors (max-pool, mean+std+max) are uniformly weak, and every other strategy is
near-flat in $k$.}}
  \label{fig:pu_heatmap}
\end{figure}

{The decoder is discarded during inference. The method used to pool the token-sequence embedding into a fixed-length descriptor and the specific $k$-NN configuration therefore materially affect detection performance. Eight pooling strategies were evaluated for each trained model. Each strategy was swept across $k\in\{1,5,10,20,50,100\}$ and the distance metric $\in\{\ell_2,\mathrm{cosine}\}$. Table~\ref{tab:pu_strategy} ranks these strategies by their mean AUROC over the 45 models with each evaluated at its respective optimal $k$ and metric combination. Figure~\ref{fig:pu_heatmap} shows the complete strategy$\times k$ grid, from which this ranking follows as the per-row optimum, and also shows that AUROC is nearly flat in $k$ for every usable strategy.}

{Three findings follow. The original $384$-dimensional \textit{mean+std+max} descriptor and plain \textit{max-pool} are the weakest usable strategies at $\approx 0.69$, because the maximum component is dominated by transient peaks on Paderborn, which motivates moment-based and concatenated descriptors. The best are the cause-effect concatenation at $0.8912$ and the cause-conditioned effect mean at $0.8846$, both stable across folds at $\mathrm{std}\approx0.045$ against the \textit{std}-pool's $\mathrm{std}\approx0.109$, whose spread comes almost entirely from the single hard fold {diagnosed in Section~\ref{subsec:pu_main}}. The AUROC is nearly insensitive to $k$, within $0.005$ across $k\in[1,100]$ for every usable strategy, unlike the hydraulic dataset where a specific $k$ is required, which indicates a compact well-separated healthy manifold on which a single neighbour suffices.}

\subsubsection{{The Stop-Gradient and Training-Set Size}}
\label{sec:finding_stopgrad}

{Across datasets the stop-gradient looks contradictory. It adds $+0.068$ AUROC on Paderborn, subtracts $-0.056$ on ZeMA, and adds $+0.044$ {to the SWaT $k$-NN score}. The cases reconcile once training-set size and the presence of a cause-encoder reconstruction loss are read together, since the datasets vary both at once. This section works through that interaction, starting with the most sensitive dataset.}

\paragraph{{Stop-Gradient Ablation on the Paderborn Dataset}}
{The mechanism loss predicts the effect embedding from the cause embedding, and the stop-gradient on the effect target stops the two encoders co-adapting into a degenerate shared representation. An otherwise identical model trained on a common split with the stop-gradient removed (Table~\ref{tab:pu_stopgrad} in the appendix) drops the overall AUROC by $0.068$ from $0.9390$ to $0.8714$. The effect-embedding standard deviation stayed well above the collapse threshold throughout, so this is not representational collapse but the two encoders drifting toward a shared less-discriminative space. The loss is largest on artificial faults at $-0.090$ and smaller on real damage at $-0.052$, so the stop-gradient is a structural part of the causal formulation rather than a stability trick.}

\paragraph{{Stop-Gradient Ablation on the ZeMA Dataset}}
\label{subsec:stopgrad_zema}

{The same ablation on ZeMA reverses. Removing the stop-gradient raises the overall AUROC by $0.056$ and cuts the seed standard deviation six-fold from $0.0409$ to $0.0065$ (Table~\ref{tab:zema_stopgrad} in the appendix, each variant at its optimal $k$). The reversal comes from the interaction between training-set size and the cause-reconstruction weight.}

{On ZeMA $\alpha_{x}=0$ (Section~\ref{subsec:zema_beta}), so the cause encoder gets no reconstruction gradient, and the stop-gradient additionally severs the mechanism gradient into the effect target, leaving the cause encoder with only the mechanism prediction error. On just $415$ cycles this starves it and its representation under-trains. Removing the stop-gradient restores a direct mechanism signal to the cause encoder, which acts as useful regularisation on a tiny dataset, unlike the harmful co-adaptation it causes on a large dataset that already supplies a reconstruction constraint such as Paderborn where $\alpha_{x}>0$.}

{Training-set size and cause-reconstruction weight vary together between Paderborn and ZeMA, so these two datasets alone leave the account under-determined and a third dataset is needed to separate the variables.}

\paragraph{{Stop-Gradient Ablation on the SWaT Dataset}}
\label{subsec:stopgrad_swat}

\begin{figure}[t]
  \centering
  \includegraphics[width=0.9982\linewidth]{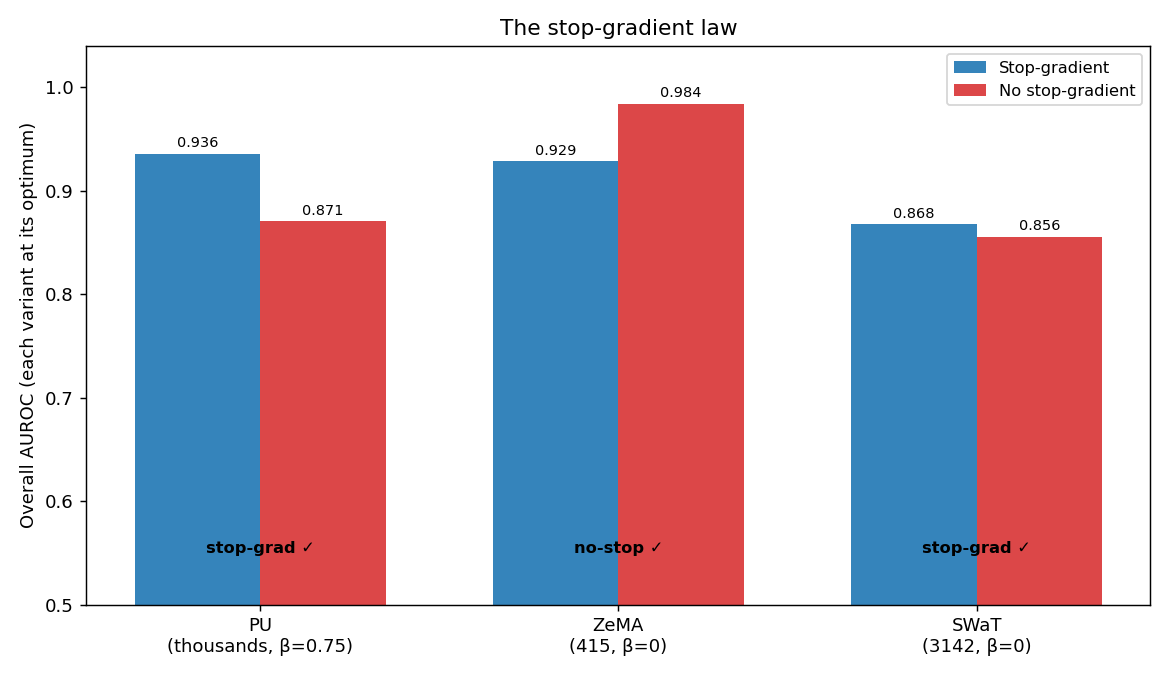}
  \caption{Stop-gradient effect across the three datasets.
  }
  \label{fig:synth_stopgrad}
\end{figure}

{On SWaT the stop-gradient variant wins, with the mechanism residual at $0.8679$ against $0.8557$ and the $k$-NN score at $0.8629$ against $0.8186$ (Figure~\ref{fig:synth_stopgrad}), matching Paderborn and opposing ZeMA. SWaT shares ZeMA's $\alpha_{x}=0$ but has enough data to avoid gradient starvation, so dataset size, not the reconstruction weight, is the primary determinant.}

{The practical rule follows. Keep the stop-gradient when the cause channel has its own reconstruction weight or when the training set is large enough to train the cause encoder through the mechanism loss alone, and omit it only when both are absent, that is an unreconstructed cause encoder on a few hundred samples.}

\subsubsection{{Manifold Geometry Across Domains}}
\label{subsec:zema_ksweep}

Rather than treating the neighbourhood size $k$ merely as a free hyperparameter to be tuned, it is more accurate to view it as a direct readout of the healthy manifold's compactness. Across the three datasets, this metric reveals three distinct geometric profiles.

ZeMA shows a pronounced optimum, contrasting sharply with Paderborn's near-indifference to $k$ (Table~\ref{tab:zema_ksweep} in the appendix, both variants). The no-stop-gradient variant peaks at $k=20$, while the stop-gradient variant requires $k=50$. This optimal value acts as a gauge for manifold quality: a compact, well-separated manifold requires only a few neighbours, whereas a noisier one relies on a wider radius to average out local irregularities. Consequently, the no-stop-gradient training produces a much better-structured space, perfectly aligning with its higher and more stable AUROC. Paderborn's indifference across $k\in[1,100]$ against ZeMA's sharp optimum reflects the fact that ZeMA's manifold, learned from only $415$ cycles around a single operating point, is more sharply defined, making the exact choice of $k$ critical. However, if $k$ is pushed too high, performance across both variants collapses to an AUROC of $\approx 0.38$ at $k=200$. As $k$ approaches the total reference bank size of $415$, the distance metric stops discriminating---a generic large-$k$ artefact observed across all datasets.

SWaT occupies the opposite geometric extreme (Table~\ref{tab:swat_cmr} in the appendix, showing the CMR $k$-sweep with the mechanism residual for both variants). Three key observations emerge here. First, the mechanism residual proves to be the strongest CMR metric, achieving $0.8679$ for the stop-gradient variant. This outperforms its own $k$-NN score and, as discussed in Section~\ref{sec:finding_scoring}, beats every baseline except the VAE. This indicates that the causal signal captures critical fault information that a simple manifold distance misses. Second, SWaT's $k$-NN score is nearly invariant to changes in $k$, hovering tightly between $0.846$ and $0.863$ across the entire $k\in[1,200]$ range. This mirrors Paderborn's stability while directly contrasting with ZeMA. Finally, SWaT suffers no large-$k$ collapse. While ZeMA crashes to $\approx 0.38$ at $k=200$, SWaT holds steady near $0.85$. This resilience proves that the cyber-attack signal is broadly distributed throughout the latent space, rather than being confined to a narrow local neighbour.

\subsection{{Cross-Domain Transfer}}
\label{sec:transfer}

{The preceding subsections evaluated all three datasets simultaneously.} {This subsection analyses}{ the specific framework modifications required to enable that cross-domain functionality. The architectural components remain largely invariant. The encoder and the loss function and the optimiser alongside the scoring rule are identical across every domain as documented in Table~\ref{tab:datasets} and Table~\ref{tab:architecture} and Table~\ref{tab:training_inference}. The framework exclusively adapts the treatment of the cause channel. This specific adaptation follows established physical principles rather than relying on empirical hyperparameter searches.} {{The comparison against the strongest baseline and the limits of the approach are taken up in the discussion of Section~\ref{sec_chap6:discussion}.}}

{The full per-domain instantiation, the causal pairs, window lengths, channel pruning, and train, validation, and test splits, is specified in Section~\ref{subsec_chap6:data_preprocessing} and Table~\ref{tab:datasets}, and is not repeated here. The one setting that does not transfer uniformly across domains is the cause-reconstruction weight, examined next.}

\subsubsection{{Cause Reconstruction Weight Across Domains}}
\label{subsec:zema_beta}

{The cause reconstruction weight represents the singular architectural setting that fails to transfer uniformly across datasets. This specific parameter requires a distinct optimal value within each evaluation domain. These numerical differences track the physical composition of the underlying cause signal rather than reflecting an arbitrary limitation of the model architecture.}

{The weight parameter $\alpha_{x}$ dictates the degree to which the cause encoder anchors to reconstructing the EPS1 signal compared to being shaped exclusively by the causal-mechanism objective. This coefficient was systematically swept across the specific values of $0.0$ and $0.25$ and $0.5$ and $0.75$ and $1.0$. Table~\ref{tab:zema_beta}
% and Figure~\ref{fig:zema_beta} 
documents the resulting performance metrics.}

{The optimum falls at $\alpha_{x}=0$, giving $0.9977$ overall. This is physically interpretable. Paderborn motor current is dominated by the electromagnetic drive that causes the vibration, so reconstructing it retains causal signal and $\alpha_{x}=0.75$ helps, whereas ZeMA motor power carries electrical overhead unrelated to the hydraulic response, so reconstructing it forces the encoder to keep noise that pollutes the predictor. At $\alpha_{x}=0$ the cause encoder acts as a pure causal feature extractor, coherent with the no-stop-gradient training of Section~\ref{sec:finding_stopgrad} in which the mechanism loss alone trains it. The sweep used a single seed, and the $\alpha_{x}=0$ choice is confirmed by the multi-seed $0.9843 \pm 0.0065$ of Section~\ref{subsec:zema_main}, with the apparent dip at $\alpha_{x}=0.5$ a seed artefact that re-evaluation places near $0.92$.}

{SWaT behaves differently. Sweeping the cause weight across $0$, $0.25$, $0.5$, $0.75$, and $1.0$ (Table~\ref{tab:swat_beta}) leaves both the mechanism residual and the $k$-NN AUROC flat, within $0.005$ across the range, so $\alpha_{x}=0$ is adopted for efficiency, since dropping the cause reconstruction removes the actuator decoders and cuts parameters. Where ZeMA motor power carries non-causal overhead that an enforced reconstruction term degrades, the heavily pruned SWaT actuator set is already a clean cause group, so adding a reconstruction weight changes nothing.}

{The SWaT dataset deviates structurally from the single-coupling bearing and hydraulic evaluations. The water treatment facility incorporates six distinct operational stages with cyber-attacks distributed broadly across all segments. A single cause-effect pair cannot adequately monitor the entire plant. The designated cause and effect groups are consequently formulated as multi-channel tensors. The empirical data strictly dictates which specific sensors carry a learnable physical mechanism. A simple linear least-squares regression model was fitted exclusively on normal operation data to evaluate this property. This regression model predicts the output of each individual sensor based entirely on the $12$ active actuator signals. The coefficient of determination was subsequently extracted to quantify the fraction of sensor variance mathematically explained by the actuators. An $R^2$ value approaching $1$ indicates that the specific sensor is fully determined by the active control commands. An $R^2$ value approaching $0$ indicates that the sensor is driven primarily by external unmonitored phenomena. This standard goodness-of-fit metric functions strictly as an initial screening statistic. The fluid flows and tank levels driven directly by mechanical pumps and valves prove to be highly predictable. The $R^2$ values span from $0.94$ to $0.99$ for the specific stage flows and reach $0.79$ for the primary stage-one tank level. The chemical analysers and reverse-osmosis pressures yield $R^2$ values below $0.15$ because they are governed entirely by complex membrane and chemistry dynamics. The mechanism residual is therefore measured exclusively across the eight strongly coupled hydraulic sensors. The comprehensive $k$-NN anomaly score continues to utilise all twenty-five sensors to ensure total plant coverage. This strict data-driven separation encodes the entire system for global anomaly detection but isolates the precise mechanism residual strictly to the channels where a provable physical coupling exists. This targeted calculation ensures the mechanism residual avoids mathematical dilution from continuous channels that lack a learnable correlation.}

\paragraph{{Optimal Cause Regularisation and Sensor Physics}}
{The architectural divergence between a $\alpha_{x}=0$ optimum on the ZeMA dataset and an $\alpha_{x}=0.75$ optimum on the Paderborn dataset represents a direct mathematical reflection of the signal-to-causal-noise ratio inherent to each cause channel rather than an arbitrary tuning artefact. Motor current functions as a predominantly causal input for mechanical bearing vibration. Motor power inherently carries non-causal electrical overhead that actively dilutes the causal signal if mathematically reconstructed. The optimal configuration of the proposed framework successfully tracks this fundamental physical distinction. This adaptability provides compelling evidence that the architecture successfully learns the intended causal relationship instead of defaulting to a generic unconstrained representation.}

\section{{Discussion and Limitations}}
\label{sec_chap6:discussion}

{The discussion treats the Paderborn results first, since they carry the full ablation, then turns to the cross-domain comparison against the strongest baseline.}

\subsection{{Discussion of the Paderborn Results}}
\label{subsec:pu_discussion}

{The Paderborn experiments establish four primary architectural and empirical conclusions.} {First, on Paderborn the framework is robust to its loss weights, though the neighbourhood size, distance metric, cause-reconstruction weight, and stop-gradient variant are each set per domain rather than universally fixed.} {The mechanism weight peaks at $\gamma=0.2$ but varies the overall AUROC by less than $0.021$ across an order-of-magnitude evaluation range. The current weight remains stable between $0.25$ and $1.0$ with a clear optimum at $0.75$. Both channels require a reconstruction anchor because setting $\alpha_{x}=0$ or setting $\alpha_{y}=0$ degrades overall performance. Second the stop-gradient operator contributes a structural improvement of $+0.068$ AUROC without inducing representation collapse.} {On Paderborn this makes it a valuable component, though Section~\ref{sec:finding_stopgrad} shows its benefit is dataset-dependent rather than universal.} {Third the anomaly scoring strategy dominates detection performance for reconstruction-based methods as detailed in Section~\ref{sec:finding_scoring}. {Under a matched fixed protocol the causal embedding of CMR-Mamba is competitive with the strongest $k$-NN baseline, trailing USAD by $0.005$ on the aggregate while leading it on the artificial-defect subset,} while remaining computationally tractable where attention-based detectors fail due to hardware limits. {Fourth CMR-Mamba achieves an overall AUROC of $0.880$ under a rigorous 15-fold leave-bearing-out cross-validation with a single fixed configuration, reaching $0.944$ on real damage with run-to-run variance bounded within $0.003$, and a per-fold oracle upper bound of $0.930$.} Total performance is limited primarily by a single hard bearing pair whose healthy signature intrinsically overlaps the faulty manifold. These collective results validate the causal-mechanism formulation on electromechanical bearing faults and establish the baseline operating parameters utilised for subsequent evaluations on the hydraulic and cyber-physical datasets.}

\subsection{{Cross-Domain Comparison and the Strongest Baseline}}
\label{subsec:synth_vae}

{The strongest baseline is a multi-seed variational autoencoder with $k$-NN scoring. It exceeds CMR-Mamba on ZeMA at $0.998$ against $0.984$ and ties it on SWaT at $0.871$ against $0.868$, within overlapping intervals. We keep it as the strongest comparator because it corroborates rather than contradicts Section~\ref{sec:finding_scoring}, since its strength comes from the $k$-NN protocol and not the variational objective, its native reconstruction score reaching only $0.66$ to $0.77$. A strong variational autoencoder under $k$-NN scoring is direct evidence that the scoring rule dominates the encoder.}

{The aggregate parity hides a decisive subset result. The variational autoencoder edges CMR-Mamba only where anomalies are blunt, on ZeMA severe leakage and the SWaT aggregate, within overlapping intervals, and the order reverses on the hard low-separability faults. The mechanism residual leads the SWaT stealthy attacks at $0.747$ while the variational autoencoder falls to $0.701$ and pure-marginal scores drop below chance, and it leads the Paderborn artificial defects} {at $0.792$ against $0.754$ under the fixed 15-fold protocol, and at $0.900$ against {the variational autoencoder's} $0.793$ on the single tuning split.} {A variational autoencoder reports only that a window is off-manifold, whereas the residual identifies which cause-to-effect relationship broke, a fault-localising signal the baseline lacks \citep{scholkopf2021causal}. The slight aggregate edge is AUROC saturation on blunt faults, not a limit of the causal architecture, which leads on the coupling-violating faults it was designed to detect.}

\paragraph{{Competitiveness across domains:}}
{A properly tuned variational autoencoder marginally exceeds CMR-Mamba on the ZeMA dataset. CMR-Mamba conversely exceeds the variational autoencoder on the Paderborn dataset, {by $0.931$ against $0.906$ under per-fold selection and by $0.8803$ against $0.8736$ under the fixed protocol, although USAD edges it on the Paderborn aggregate under that same fixed protocol.} {No single baseline model is simultaneously the strongest across all three datasets.} {USAD's aggregate edge is specific to Paderborn and does not generalise: it trails CMR-Mamba by $0.0329$ on ZeMA ($0.9514$ against $0.9843$) and by $0.0020$ on SWaT ($0.8659$ against $0.8679$), so it is not the cross-domain competitor that the variational autoencoder is.} The standard autoencoder collapses on ZeMA while the Anomaly Transformer remains computationally intractable on Paderborn and the variational autoencoder trails on Paderborn. CMR-Mamba consistently performs within $0.014$ of the best method on every evaluated domain. It additionally provides a causal and interpretable representation alongside a mechanism-residual signal that the reconstruction baselines entirely lack. This core contribution is therefore best defined as consistent cross-domain competitiveness paired with structural interpretability rather than as an isolated per-dataset leaderboard victory.}

\section{Conclusion}
\label{sec:conclusion}

This research introduced CMR-Mamba as an unsupervised fault detector that monitors the causal coupling between distinct sensor groups rather than evaluating the marginal appearance of individual signals. The framework pairs per-domain Mamba state-space encoders with a causal cross-modal predictor. It scores anomalies utilising a $k$-nearest neighbour distance calculated on the healthy effect manifold alongside a dedicated mechanism residual. The architecture was evaluated across three physically unrelated coupling-fault domains comprising electromechanical bearings and hydraulic leakage and a cyber-physical water-treatment testbed. CMR-Mamba proves competitive with the strongest baseline models on the aggregate metrics across all environments.  On the aggregate the leading methods are statistically indistinguishable, and a paired Wilcoxon test on the Paderborn folds cannot separate CMR-Mamba from USAD, the VAE, or the autoencoder, while the differences on the hydraulic and cyber-physical datasets fall within overlapping intervals, so no single method dominates the aggregate. CMR-Mamba is nonetheless the best of the evaluated learned detectors on the low-separability hard subset of every dataset that has one, reaching $0.792$ against $0.754$ on the Paderborn artificial defects and $0.747$ against $0.701$ on the SWaT stealthy attacks, where marginal methods sit at chance. {I}t achieves this performance while supplying an interpretable and fault-localizing signal that pure reconstruction methods cannot mathematically provide. Two primary findings generalise well beyond this specific architecture. The anomaly-scoring strategy rather than the chosen encoder family operates as the dominant driver of detection performance. The $k$-NN manifold scoring protocol improves every reconstruction baseline substantially more than any structural change to the encoder. The mechanism residual provides a distinct and highly security-relevant detection capability. It successfully isolates stealthy cyber-physical attacks that keep every individual sensor within its normal historical range to evade standard marginal detectors. The scope of the method is stated plainly. It targets coupling faults, and a fault that leaves the instantaneous cause-to-effect relationship intact is by construction invisible to it, whatever score is applied. Fault classes of that kind were not evaluated here. Future work will include comprehensive per-domain parameter sweeps of the cause-reconstruction weight to definitively confirm its transferability. Future research will also encompass cross-rig validation of the hydraulic findings and the mathematical extension of the mechanism residual to achieve multi-stage root-cause localization. The ultimate objective remains the integration of this structured perception module with normative reward-learning frameworks to realize fully autonomous industrial maintenance agents.

\onecolumn
\section*{{Appendix}}
\label{sec:chap6_appendix}

\subsection*{{Full Architecture and Training Specification}}
\label{sec:full_spec}

{This appendix gives the complete per-domain specification summarised in the methodology. Table~\ref{tab:training_inference} lists the full training and inference settings, and Table~\ref{tab:architecture} lists the complete architecture. {The main text} reports only the settings needed to read the results, and the full detail is collected here for reproducibility.}

\begingroup
\renewcommand{\arraystretch}{1.15}
\small
\setlength{\tabcolsep}{4pt}
\begin{xltabular}{\textwidth}{@{}
    >{\raggedright\arraybackslash\hsize=1.5\hsize}X
    >{\raggedright\arraybackslash\hsize=0.9\hsize}X
    >{\raggedright\arraybackslash\hsize=0.75\hsize}X
    >{\raggedright\arraybackslash\hsize=0.85\hsize}X @{}}
\caption{Training and inference settings. The optimiser, gradient handling, and metric are common to all three domains. The loss weights and the inference settings differ, and they follow the physical asymmetry in cause-channel signal-to-noise ratio discussed in Section~\ref{subsec_chap6:causal_training}.}
\label{tab:training_inference}
\\
\toprule
\textbf{Hyperparameter} & \textbf{Paderborn} & \textbf{ZeMA} & \textbf{SWaT} \\
\midrule
\endfirsthead
\multicolumn{4}{@{}l}{\small\itshape Table~\ref{tab:training_inference} continued from previous page}\\
\toprule
\textbf{Hyperparameter} & \textbf{Paderborn} & \textbf{ZeMA} & \textbf{SWaT} \\
\midrule
\endhead
\midrule
\multicolumn{4}{r@{}}{\small\itshape continued on next page}\\
\endfoot
\bottomrule
\endlastfoot
\multicolumn{4}{@{}l}{\textit{Loss weights (Eqs.~\ref{eq:loss-mech}--\ref{eq:loss-recon-c})}} \\
$\gamma$ (mechanism) & 0.20 & 0.20 & 0.20 \\
$\alpha_{y}$ (effect reconstruction) & 1.00 & 1.00 & 1.00 \\
$\alpha_{x}$ (cause reconstruction) & 0.75 & 0.00 & 0.00 \\
\midrule
\multicolumn{4}{@{}l}{\textit{Optimiser and schedule}} \\
Optimiser & AdamW & AdamW & AdamW \\
Initial learning rate & $5 \times 10^{-4}$ & $5 \times 10^{-4}$ & $5 \times 10^{-4}$ \\
Weight decay & $10^{-5}$ & $10^{-5}$ & $10^{-5}$ \\
$\beta_1$, $\beta_2$ & 0.9, 0.999 & 0.9, 0.999 & 0.9, 0.999 \\
Warm-up & 5 epochs (linear) & 5 epochs (linear) & 5 epochs (linear) \\
Post-warm-up schedule & cosine annealing & ReduceLROnPlateau & ReduceLROnPlateau \\
Gradient clipping norm & 1.0 & 1.0 & 1.0 \\
\midrule
\multicolumn{4}{@{}l}{\textit{Training budget}} \\
Batch size & 12 & 32 & 32 \\
Maximum epochs & 100 & 200 & 150 \\
Early-stopping patience & 10 epochs & 15 epochs & 15 epochs \\
Training set size & $\sim$4,500 windows & 415 cycles & 3,142 windows \\
Wall-clock per fold or seed & $\sim$3 hours & $\sim$20 minutes & $\sim$3.3 hours \\
Hardware & \multicolumn{3}{@{}l@{}}{Shared cluster (RTX A4000, RTX 4000/4500 Ada, L40S, A100 40\,GB)} \\
\midrule
\multicolumn{4}{@{}l}{\textit{Inference}} \\
Reported score & $k$-NN distance & $k$-NN distance & mechanism residual, $k$-NN \\
Temporal pooling $\phi$ & std (Eq.~\ref{eq:pool}) & std (Eq.~\ref{eq:pool}) & std (Eq.~\ref{eq:pool}) \\
Neighbourhood size $k$ & 1 & 20 & 5 \\
Distance metric & cosine & $L_2$ & $L_2$ \\
Reference bank & all training windows & 415 healthy cycles & 3,142 healthy windows \\
\midrule
\multicolumn{4}{@{}l}{\textit{Evaluation}} \\
Validation strategy & leave-two-out & temporal split & normal/attack split \\
Runs reported & 15 ($\binom{6}{2}$ pairs) & 3 seeds & 3 seeds \\
Random seeds & deterministic & 1337, 9999, 777 & 1337, 9999, 777 \\
Primary metric & AUROC & AUROC & AUROC, point-wise F1 \\
Stratified metric & artificial vs.\ real & weak vs.\ severe & stealthy vs.\ blunt \\
\end{xltabular}
\endgroup

\begingroup
\renewcommand{\arraystretch}{1.15}
\small
\setlength{\tabcolsep}{4pt}
\begin{xltabular}{\textwidth}{@{}
    >{\raggedright\arraybackslash\hsize=1.5\hsize}X
    >{\raggedright\arraybackslash\hsize=0.9\hsize}X
    >{\raggedright\arraybackslash\hsize=0.75\hsize}X
    >{\raggedright\arraybackslash\hsize=0.85\hsize}X @{}}
\caption{Complete architecture specification for the three dataset instantiations. The PerChannelMambaEncoder hyperparameters are identical across datasets{, and} the only architectural differences sit at the input boundary and at the encoder/decoder sharing pattern, and they arise mechanically from the different sampling rates and sensor counts of the three acquisition systems.}
\label{tab:architecture}
\\
\toprule
\textbf{Hyperparameter} & \textbf{Paderborn} & \textbf{ZeMA} & \textbf{SWaT} \\
\midrule
\endfirsthead
\multicolumn{4}{@{}l}{\small\itshape Table~\ref{tab:architecture} continued from previous page}\\
\toprule
\textbf{Hyperparameter} & \textbf{Paderborn} & \textbf{ZeMA} & \textbf{SWaT} \\
\midrule
\endhead
\midrule
\multicolumn{4}{r@{}}{\small\itshape continued on next page}\\
\endfoot
\bottomrule
\endlastfoot
\multicolumn{4}{@{}l}{\textit{Input}} \\
Sampling rate & 64\,kHz & 100\,Hz & 1\,Hz \\
Input window length (samples) & 32,768 & 6,000 & 256 \\
Sensor channels per window & 3 & 2 & 37 (12 cause + 25 effect) \\
Cause channel(s) & Current phase A, B & EPS1 (motor power) & 12 active actuators (pumps, valves) \\
Effect channel & Vibration & PS1 (pump pressure) & 25 process sensors \\
\midrule
\multicolumn{4}{@{}l}{\textit{ConvStem (one per encoder instance)}} \\
Input $\to$ output channels & $1 \to 128$ & $1 \to 128$ & $1 \to 128$ \\
Kernel size & 7 & 7 & 7 \\
Stride & 4 & 4 & 4 \\
Padding & 3 & 3 & 3 \\
Output token sequence length $L'$ & 8,192 & 1,500 & 64 \\
Post-convolution normalisation & BatchNorm1d & BatchNorm1d & BatchNorm1d \\
Post-convolution activation & GELU & GELU & GELU \\
\midrule
\multicolumn{4}{@{}l}{\textit{Mamba block (each of four stacked)}} \\
Number of stacked blocks $N_{\mathrm{enc}}$ & 4 & 4 & 4 \\
Model dimension $d_{\mathrm{model}}$ & 128 & 128 & 128 \\
State dimension $d_{\mathrm{state}}$ & 16 & 16 & 16 \\
Depthwise convolution width $d_{\mathrm{conv}}$ & 4 & 4 & 4 \\
Expansion factor & 2 & 2 & 2 \\
Internal hidden dimension after expansion & 256 & 256 & 256 \\
Normalisation placement & pre-norm & pre-norm & pre-norm \\
Final LayerNorm after the stack & yes & yes & yes \\
\midrule
\multicolumn{4}{@{}l}{\textit{Encoder instances per dataset}} \\
Number of encoder instances & 2 & 2 & 2 \\
Effect-channel encoder & \textit{enc\_vib} (dedicated) & \textit{enc\_ps1} (dedicated) & \textit{enc\_effect} (shared on 25 sensors) \\
Cause-channel encoder & \textit{enc\_curr} (shared on A, B) & \textit{enc\_eps1} (dedicated) & \textit{enc\_cause} (shared on 12 actuators) \\
\midrule
\multicolumn{4}{@{}l}{\textit{Causal predictor}} \\
Fusion input dimension & 256 (concatenated A, B) & 128 (single channel) & 1,536 ($12\times128$) \\
Fusion output dimension & 128 & 128 & 3,200 ($25\times128$) \\
Fusion dropout probability & 0.1 & 0.1 & 0.1 \\
Number of causal convolutional blocks & 3 & 3 & 3 \\
Convolution kernel size & 5 & 5 & 5 \\
Padding mode & left-only (causal) & left-only (causal) & left-only (causal) \\
Receptive field (tokens) & 13 & 13 & 13 \\
Receptive field (physical time) & 0.81\,ms & 520\,ms & 52\,s \\
Output projection initialisation & identity & identity & default \\
\midrule
\multicolumn{4}{@{}l}{\textit{Decoder (one per modality)}} \\
Stage 1: feature-enrichment convolution & $128 \to 128$ channels & $128 \to 128$ channels & $128 \to 128$ channels \\
Stage 2: ConvTranspose1d ($\times 2$ upsample) & $128 \to 64$ channels & $128 \to 64$ channels & $128 \to 64$ channels \\
Stage 3: ConvTranspose1d ($\times 2$ upsample) & $64 \to 32$ channels & $64 \to 32$ channels & $64 \to 32$ channels \\
Stage 4: projection convolution ($k{=}7, p{=}3$) & $32 \to 1$ channel & $32 \to 1$ channel & $32 \to 1$ channel \\
Output waveform length & 32,768 & 6,000 & 256 \\
Per-stage normalisation / activation & BatchNorm1d / GELU & BatchNorm1d / GELU & BatchNorm1d / GELU \\
Decoder sharing & cause decoder shared on {A, B} & none & effect decoder per sensor{, and} cause decoder omitted ({$\alpha_{x}{=}0$}) \\
\midrule
\multicolumn{4}{@{}l}{\textit{Total model size}} \\
Trainable parameters & 1.67\,M & 1.59\,M & 5.41\,M \\
\end{xltabular}
\endgroup

\subsection*{{Paderborn Bearing Subset}}
\label{sec:pu_bearing_subset}

This table lists the 25 Paderborn bearing experiments used in the study and the reason each faulty bearing was selected. {The main text} refers to it for the composition of the evaluation set.

\begingroup
\renewcommand{\arraystretch}{1.15}
\small
\begin{xltabular}{0.75\textwidth}{lccc X}
\caption{The 25 PU bearing experiments used in this study: 6 healthy bearings used for both training and held-out testing under leave-bearings-out cross-validation, and 19 faulty bearings for evaluation. OR = outer race; IR = inner race; IR+OR = combined.}
\label{tab:pu_bearings}
\\
\toprule
\textbf{Bearing} & \textbf{Damage Type} & \textbf{Location} & \textbf{Severity} & \textbf{Selection Rationale} \\
\midrule
\endfirsthead
\multicolumn{5}{@{}l}{\small\itshape Table~\ref{tab:pu_bearings} continued from previous page}\\
\toprule
\textbf{Bearing} & \textbf{Damage Type} & \textbf{Location} & \textbf{Severity} & \textbf{Selection Rationale} \\
\midrule
\endhead
\midrule
\multicolumn{5}{r@{}}{\small\itshape continued on next page}\\
\endfoot
\bottomrule
\endlastfoot
\multicolumn{5}{l}{\textit{Healthy Bearings (no fault; used for training and held-out test under cross-validation)}} \\
\midrule
K001--K006 & --- & --- & --- & Full healthy envelope \\
\midrule
\multicolumn{5}{l}{\textit{Artificial Faults (evaluation / test only)}} \\
\midrule
KA01 & EDM & OR & 1 & Sole OR EDM Level~1 \\
KA03 & Engraver & OR & 2 & OR engraver Level~2; severity pair with KA05 \\
KA05 & Engraver & OR & 1 & OR engraver Level~1; high mechanism-shift potential \\
KA07 & Drilling & OR & 1 & OR drilling Level~1; hard fault for CMR validation \\
KA08 & Drilling & OR & 2 & OR drilling Level~2; severity pair with KA07 \\
KI01 & EDM & IR & 1 & Sole IR EDM Level~1 \\
KI03 & Engraver & IR & 1 & IR engraver Level~1; severity pair with KI07 \\
KI07 & Engraver & IR & 2 & IR engraver Level~2; moderate difficulty \\
\midrule
\multicolumn{5}{l}{\textit{Real (Lifetime) Faults (evaluation / test only)}} \\
\midrule
KA04 & Fatigue pitting & OR & 1 & OR pitting Level~1; high mechanism-shift potential \\
KA15 & Plastic deform. & OR & 1 & Sole OR plastic deformation specimen \\
KA16 & Fatigue pitting & OR & 2 & OR pitting Level~2; severity pair with KA04/KA22 \\
KA22 & Fatigue pitting & OR & 1 & OR pitting Level~1; hardest real OR fault~\cite{neupane2025multisensor}\\
KA30 & Plastic deform & OR & 1 & Distributed OR plastic deformation \\
KB23 & Fatigue pitting & IR+OR & 2 & Combined fault Level~2; compound coupling disruption \\
KB24 & Fatigue pitting & IR+OR & 3 & Combined fault Level~3; severity pair with KB23 \\
KB27 & Plastic deform. & IR+OR & 1 & Sole combined plastic deformation specimen \\
KI04 & Fatigue pitting & IR & 1 & IR pitting Level~1; hardest fault in dataset ~\cite{neupane2025multisensor} \\
KI14 & Fatigue pitting & IR & 1 & IR pitting Level~1; second hard IR specimen \\
KI16 & Fatigue pitting & IR & 2 & IR pitting Level~2; severity pair with KI04/KI14 \\
\end{xltabular}
\endgroup

\subsection*{{Per-Fold and Per-Seed Detail}}
\label{sec:perfold_detail}

Table~\ref{tab:pu_main} gives the per-fold Paderborn AUROC and Table~\ref{tab:zema_main} the per-seed ZeMA AUROC. {The main text reports their aggregates, and} the full breakdowns are collected here.

\begin{table}[H] 
  \centering
  \caption{Per-fold AUROC on PU (3-run mean), under
  $(\alpha_{x},\alpha_{y},\gamma)=(0.75,1.0,0.2)$. Each
  fold holds out the listed healthy bearing pair.}
  \label{tab:pu_main}
  \begin{tabular}{clccccc}
    \toprule
    Split & Test bearings & run1 & run2 & run3 & \textbf{Mean} & Std \\
    \midrule
    0 & K001,K002 & 0.9303 & 0.9303 & 0.9390 & 0.9332 & 0.0041 \\
    1 & K001,K003 & 1.0000 & 0.9998 & 0.9996 & \textbf{0.9998} & 0.0002 \\
    2 & K001,K004 & 0.9424 & 0.9515 & 0.9543 & 0.9494 & 0.0051 \\
    3 & K001,K005 & 0.9429 & 0.9475 & 0.9530 & 0.9478 & 0.0041 \\
    4 & K001,K006 & 0.8808 & 0.8923 & 0.8820 & 0.8850 & 0.0052 \\
    5 & K002,K003 & 0.9220 & 0.8994 & 0.9145 & 0.9120 & 0.0094 \\
    6 & K002,K004 & 0.8807 & 0.8826 & 0.8923 & 0.8852 & 0.0051 \\
    7 & K002,K005 & 0.9162 & 0.9412 & 0.9393 & 0.9322 & 0.0114 \\
    8 & K002,K006 & 0.9254 & 0.9130 & 0.8886 & 0.9090 & 0.0153 \\
    9 & K003,K004 & 0.9667 & 0.9299 & 0.9468 & 0.9478 & 0.0150 \\
    10 & K003,K005 & 0.9458 & 0.9359 & 0.9305 & 0.9374 & 0.0063 \\
    11 & K003,K006 & 0.9747 & 0.9778 & 0.9985 & 0.9837 & 0.0106 \\
    12 & K004,K005 & 0.8705 & 0.8596 & 0.8548 & \textbf{0.8616} & 0.0066 \\
    13 & K004,K006 & 0.9187 & 0.9339 & 0.9155 & 0.9227 & 0.0080 \\
    14 & K005,K006 & 0.9523 & 0.9275 & 0.9373 & 0.9390 & 0.0102 \\
    \midrule
    \multicolumn{5}{l}{\textbf{Grand mean (15 folds)}} & \textbf{0.9297} & 0.0351 \\
    \bottomrule
  \end{tabular}
\end{table}

\begin{table}[H] 
  \centering
  \caption{{Per-fold Paderborn AUROC under the fixed evaluation protocol (std-pooling, $k=1$, cosine), for CMR-Mamba (three-run mean) and the four baselines evaluated at the same matched protocol. These are the paired per-fold values behind the Wilcoxon signed-rank test reported in Section~\ref{subsec:pu_main}. CMR-Mamba exceeds USAD on nine of the fifteen folds, the VAE on nine, the autoencoder on eleven, and Deep SVDD on fourteen.}}
  \label{tab:pu_fixed_perfold}
  \begin{tabular}{lccccc}
    \toprule
    Fold & CMR-Mamba & USAD & VAE & AE & Deep SVDD \\
    \midrule
    0 & 0.8913 & 0.8922 & 0.8987 & 0.8898 & 0.6826 \\
    1 & 0.9950 & 0.9034 & 0.9122 & 0.8830 & 0.6845 \\
    2 & 0.9430 & 0.9415 & 0.9033 & 0.8646 & 0.6564 \\
    3 & 0.9418 & 0.9042 & 0.8906 & 0.8548 & 0.6361 \\
    4 & 0.8823 & 0.8153 & 0.6849 & 0.7426 & 0.6776 \\
    5 & 0.8909 & 0.8758 & 0.9018 & 0.8838 & 0.7045 \\
    6 & 0.8124 & 0.8970 & 0.8646 & 0.8501 & 0.6508 \\
    7 & 0.7316 & 0.9053 & 0.8124 & 0.8065 & 0.6149 \\
    8 & 0.8434 & 0.8770 & 0.8991 & 0.8836 & 0.6352 \\
    9 & 0.9470 & 0.9188 & 0.9191 & 0.9276 & 0.6464 \\
    10 & 0.9370 & 0.8884 & 0.9005 & 0.9176 & 0.6262 \\
    11 & 0.9827 & 0.9077 & 0.8805 & 0.8779 & 0.6636 \\
    12 & 0.5561 & 0.7371 & 0.8409 & 0.8068 & 0.7930 \\
    13 & 0.9128 & 0.9191 & 0.8859 & 0.8972 & 0.6329 \\
    14 & 0.9370 & 0.8930 & 0.9100 & 0.9253 & 0.6465 \\
    \midrule
    Mean & $0.8803 \pm 0.1085$ & $0.8850 \pm 0.0478$ & $0.8736 \pm 0.0575$ & $0.8674 \pm 0.0484$ & $0.6634 \pm 0.0420$ \\
    \bottomrule
  \end{tabular}
\end{table}

\begin{table}[H] 
  \centering
  \caption{Per-seed AUROC on ZeMA at the optimal neighbourhood size.
  No-stop-gradient at $k=20${, and} stop-gradient at $k=50$.}
  \label{tab:zema_main}
  \begin{tabular}{llccc}
    \toprule
    Variant & Seed & Overall & Weak & Severe \\
    \midrule
    \multirow{4}{*}{No stop-gradient ($k{=}20$)}
      & 1337 & 0.9897 & 0.9794 & 1.0000 \\
      & 9999 & 0.9881 & 0.9765 & 0.9997 \\
      & 777  & 0.9752 & 0.9506 & 0.9997 \\
      \cmidrule(lr){2-5}
      & \textbf{mean} & $\mathbf{0.9843 \pm 0.0065}$ & 0.9688 & 0.9998 \\
    \midrule
    \multirow{4}{*}{Stop-gradient ($k{=}50$)}
      & 1337 & 0.9084 & 0.8311 & 0.9857 \\
      & 9999 & 0.9858 & 0.9716 & 1.0000 \\
      & 777  & 0.8921 & 0.8102 & 0.9740 \\
      \cmidrule(lr){2-5}
      & mean & $0.9288 \pm 0.0409$ & 0.8710 & 0.9866 \\
    \bottomrule
  \end{tabular}
\end{table}

\subsection*{{Per-Dataset Baseline Detail}}
\label{sec:baseline_detail}

\begin{figure}[H]
  \centering
  \begin{subfigure}{\linewidth}
    \centering
    \includegraphics[height=0.285\textheight,keepaspectratio]{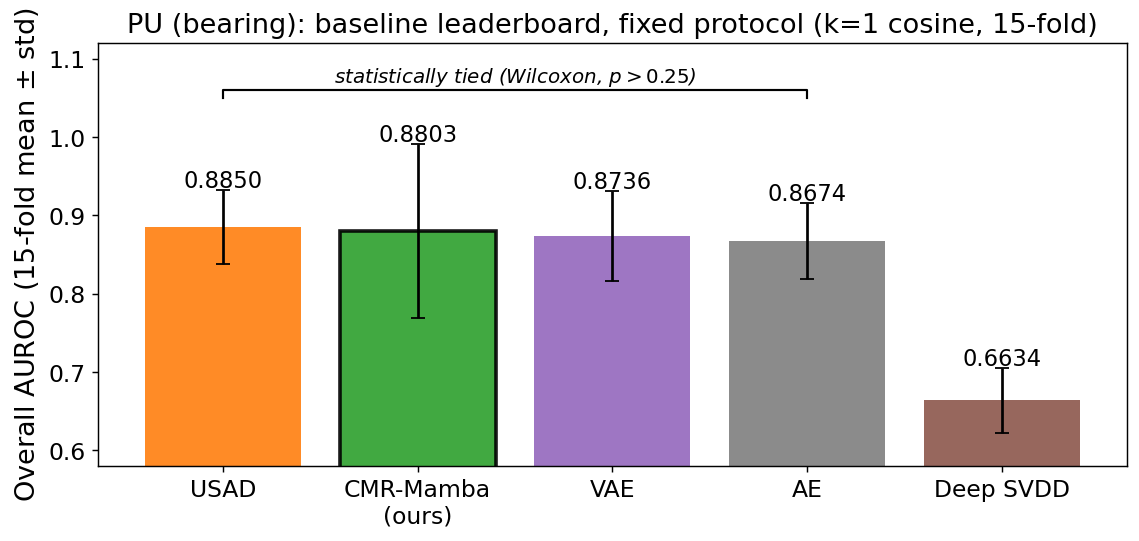}
    \caption{Paderborn.}
  \end{subfigure}\\[4pt]
  \begin{subfigure}{\linewidth}
    \centering
    \includegraphics[height=0.285\textheight,keepaspectratio]{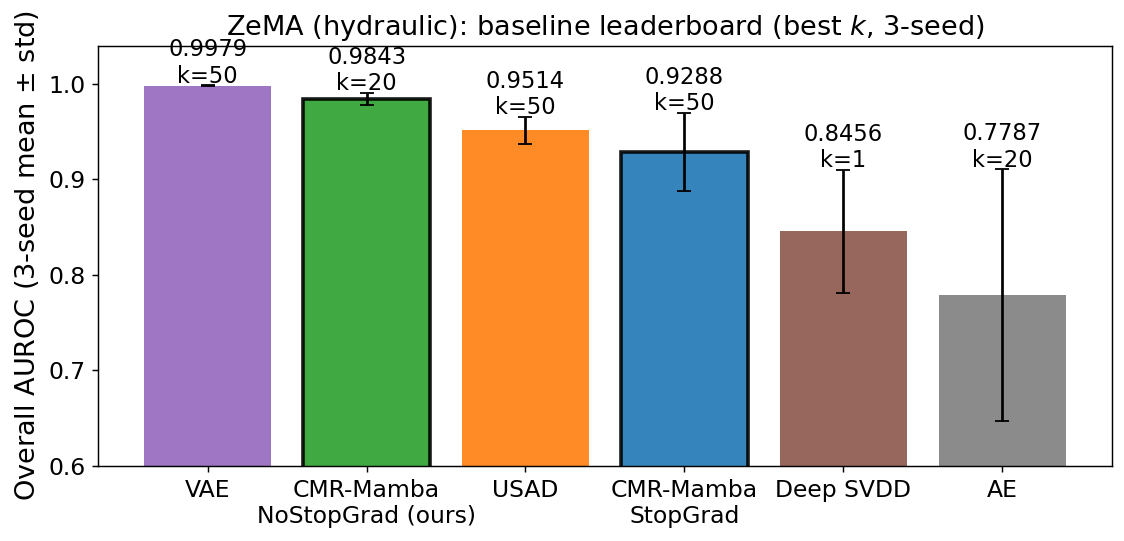}
    \caption{ZeMA.}
  \end{subfigure}\\[4pt]
  \begin{subfigure}{\linewidth}
    \centering
    \includegraphics[height=0.285\textheight,keepaspectratio]{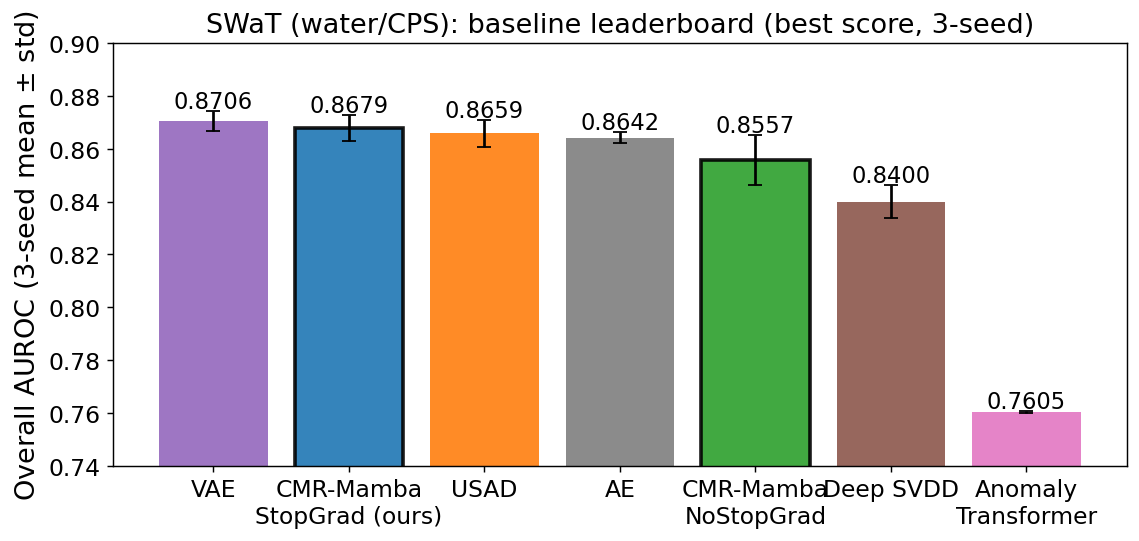}
    \caption{SWaT.}
  \end{subfigure}
  \caption{{Baseline comparison across the three domains, each panel a ranked leaderboard with the CMR-Mamba bars outlined. (a) On Paderborn the top four methods are statistically tied, as the bracket marks, and CMR-Mamba's wide error bar comes from the single hard fold in which a mutually covering bearing pair collapses class separation. (b) On ZeMA, CMR-Mamba is competitive with the VAE and far more stable than the AE. (c) On SWaT, its mechanism residual is second behind the VAE.}}
  \label{fig:baselines_all}
\end{figure}

{These tables give the per-dataset baseline comparisons in full, including native reconstruction scores, optimal neighbourhood sizes, and per-method ranks. The consolidated cross-dataset view is Table~\ref{tab:synth_master} in {the main text}.} {The Anomaly Transformer is absent from the Paderborn comparison because its $\mathcal{O}(N^2)$ attention exhausts GPU memory at that window length, leaving only 4 of the 15 folds evaluable, and the Paderborn panel of Figure~\ref{fig:baselines_all} uses the fixed evaluation protocol matched across methods.}

\begin{table}[H]
  \centering
  \caption{Baseline comparison on PU (single split). $k$-NN scoring on the same
  embeddings is reported alongside each method's native score. The Anomaly
  Transformer is intractable at the PU window length.}
  \label{tab:pu_baselines}
  \begin{tabular}{lcc}
    \toprule
    Method & $k$-NN AUROC & Native / reconstruction \\
    \midrule
    AE                  & 0.9072 & 0.8486 \;(recon MSE) \\
    VAE                 & 0.9057 & 0.6620 \;(recon MSE) \\
    USAD                & 0.8990 & 0.5367 \;(dual recon) \\
    Deep SVDD           & 0.6458 & 0.8920 \;(centre dist.) \\
    Anomaly Transformer & \multicolumn{2}{c}{OOM, intractable at $N{=}8,192$ tokens ($\mathcal{O}(N^2)$)} \\
    \midrule
    \textbf{CMR-Mamba (ours)} & \multicolumn{2}{c}{$\approx 0.955$ (this split); $0.9297\pm0.0351$ (15-fold)} \\
    \bottomrule
  \end{tabular}
\end{table}

\begin{table}[H] 
  \centering
  \caption{Baseline comparison on ZeMA (3-seed mean $\pm$ std, each method at its
  optimal $k$), with the native reconstruction score for reference. CMR-Mamba is
  competitive with the strongest baseline, and $k$-NN scoring lifts every
  reconstruction method far above its native score.}
  \label{tab:zema_baselines}
  \begin{tabular}{lccc}
    \toprule
    Method & Optimal $k$ & Overall AUROC ($k$-NN) & Native score \\
    \midrule
    VAE                       & 50 & $0.9979 \pm 0.0006$ & 0.9115 \;(recon MSE) \\
    \textbf{CMR-Mamba (NoStopGrad)} & 20 & $\mathbf{0.9843 \pm 0.0065}$ & --- \\
    USAD                      & 50 & $0.9514 \pm 0.0144$ & 0.5267 \;(dual recon) \\
    CMR-Mamba (StopGrad)      & 50 & $0.9288 \pm 0.0409$ & --- \\
    Deep SVDD                 & 1  & $0.8456 \pm 0.0646$ & 0.5641 \;(centre dist.) \\
    AE                        & 20 & $0.7787 \pm 0.1322$ & 0.3535 \;(recon MSE) \\
    \bottomrule
  \end{tabular}
\end{table}

\begin{table}[H]
  \centering
  \small
  \caption{SWaT leaderboard (3-seed mean $\pm$ std, each method at its best score).}
  \label{tab:swat_baselines}
  \begin{tabular}{clcc}
    \toprule
    Rank & Method & Best AUROC & Score \\
    \midrule
    1 & VAE & $0.8706 \pm 0.0037$ & $k$-NN ($k{=}5$) \\
    2 & \textbf{CMR-Mamba (StopGrad)} & $0.8679 \pm 0.0048$ & mechanism residual \\
    3 & USAD & $0.8659 \pm 0.0051$ & $k$-NN ($k{=}1$) \\
    4 & AE & $0.8642 \pm 0.0021$ & $k$-NN ($k{=}5$) \\
    5 & CMR-Mamba (NoStopGrad) & $0.8557 \pm 0.0094$ & mechanism residual \\
    6 & Deep SVDD & $0.8400 \pm 0.0064$ & $k$-NN ($k{=}200$) \\
    7 & Anomaly Transformer & $0.7605 \pm 0.0004$ & $k$-NN ($k{=}1$) \\
    \midrule
    \multicolumn{2}{l}{AE / VAE / USAD, native recon.} & \multicolumn{2}{c}{$\approx 0.769$} \\
    \bottomrule
  \end{tabular}
\end{table}

\subsection*{{Ablation Sweep Detail}}
{The loss-weight, mechanism-weight, and embedding-strategy sweeps referenced in Section~\ref{subsec_chap6:ablations} are collected here. {The main text reports the selected operating points, and} the full sweeps follow.}

\begin{figure}[H]
  \centering
  \includegraphics[width=0.92\linewidth]{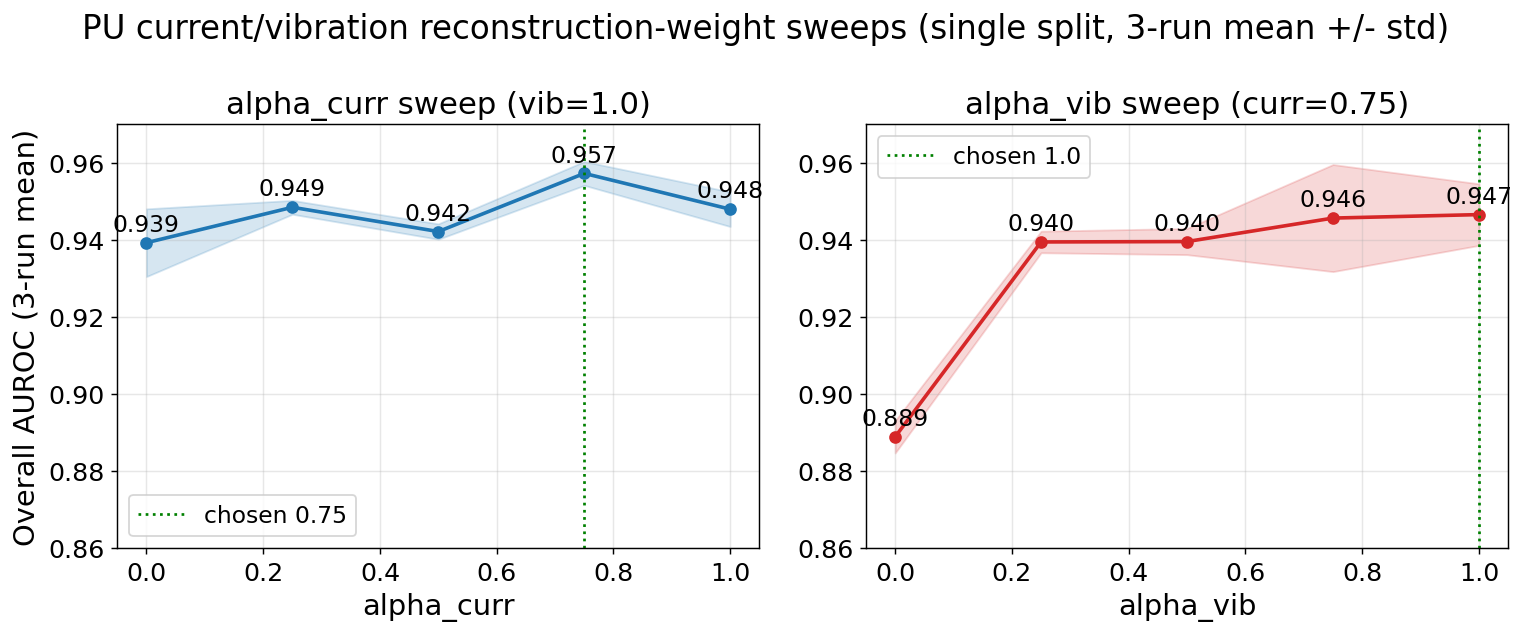}
  \caption{{Current (left) and vibration (right)
  reconstruction-weight sweeps on PU (three-run mean, shaded $\pm$ std). The
  current weight peaks at $\alpha_{x}=0.75$. The vibration weight
  degrades sharply at $\alpha_{y}=0$, where the effect encoder loses
  its reconstruction anchor, then forms a flat plateau across
  $\alpha_{y}\in\{0.5,0.75,1.0\}$ on which $\alpha_{y}=1.0$
  is adopted.}}
  \label{fig:pu_currvib}
\end{figure}

\begin{table}[H] 
  \centering
  \caption{Mechanism-weight ($\gamma$) sweep on PU (single split,
  $\alpha_{x}=0.75$, $\alpha_{y}=1.0$). Overall AUROC
  peaks at $\gamma=0.2$.}
  \label{tab:pu_gamma}
  \begin{tabular}{lccc}
    \toprule
    $\gamma$ & Overall & Artificial & Real \\
    \midrule
    0.00 & 0.9339 & 0.8554 & 0.9909 \\
    0.05 & 0.9203 & 0.8330 & 0.9838 \\
    0.10 & 0.9370 & 0.8576 & 0.9947 \\
    \textbf{0.20} & \textbf{0.9405} & \textbf{0.8739} & 0.9889 \\
    0.50 & 0.9229 & 0.8302 & 0.9903 \\
    1.00 & 0.9324 & 0.8462 & 0.9950 \\
    \bottomrule
  \end{tabular}
\end{table}

\begin{table}[H] 
  \centering
  \small
  \caption{Embedding-strategy ranking on PU (mean over 45 models, each strategy
  at its own best $k$/metric). The original \textit{mean+std+max} descriptor and
  \textit{max-pool} collapse{, while} concatenation and \textit{std}-pool dominate.}
  \label{tab:pu_strategy}
  \begin{tabular}{clccc}
    \toprule
    Rank & Strategy & Mean AUROC & Std & ($k$, metric) \\
    \midrule
    1 & vib+curr concat ($D{=}256$)      & \textbf{0.8912} & 0.045 & (100, $\ell_2$) \\
    2 & curr-enc(vib) mean ($D{=}128$)   & 0.8846 & 0.047 & (50, $\ell_2$) \\
    3 & std-pool ($D{=}128$)             & 0.8803 & 0.109 & (1, cosine) \\
    4 & mean-pool ($D{=}128$)            & 0.8661 & 0.056 & (1, cosine) \\
    5 & mean-pool $\to$ PCA(64)          & 0.8651 & 0.057 & (1, $\ell_2$) \\
    6 & mean+std ($D{=}256$)             & 0.8603 & 0.080 & (1, cosine) \\
    {7} & max-pool ($D{=}128$)             & 0.6960 & 0.122 & (100, cosine) \\
    {8} & mean+std+max ($D{=}384$)         & 0.6882 & 0.113 & (100, cosine) \\
    \bottomrule
  \end{tabular}
\end{table}

\begin{table}[H]
  \centering
  \caption{Stop-gradient ablation on PU (matched, same split). Removing the
  stop-gradient costs $0.068$ overall AUROC{, and} no representational collapse occurs.}
  \label{tab:pu_stopgrad}
  \begin{tabular}{lccc}
    \toprule
    Configuration & Overall & Artificial & Real \\
    \midrule
    With stop-gradient    & \textbf{0.9390} & 0.8779 & 0.9835 \\
    Without stop-gradient & 0.8714 & 0.7880 & 0.9319 \\
    \midrule
    $\Delta$              & $-0.0676$ & $-0.0899$ & $-0.0516$ \\
    \bottomrule
  \end{tabular}
\end{table}

\begin{table}[H] 
  \centering
  \caption{Stop-gradient comparison on ZeMA (3-seed mean $\pm$ std, each variant
  at its optimal $k$). Removing the stop-gradient improves overall AUROC by
  $0.056$ and reduces the seed variance sixfold.}
  \label{tab:zema_stopgrad}
  \begin{tabular}{lcccc}
    \toprule
    Variant & Optimal $k$ & Overall & Weak & Severe \\
    \midrule
    Stop-gradient    & 50 & $0.9288 \pm 0.0409$ & 0.8710 & 0.9866 \\
    \textbf{No stop-gradient} & 20 & $\mathbf{0.9843 \pm 0.0065}$ & 0.9688 & 0.9998 \\
    \midrule
    $\Delta$ (NoStop $-$ Stop) & & $+0.0555$ & $+0.0978$ & $+0.0132$ \\
    \bottomrule
  \end{tabular}
\end{table}

\begin{table}[H] 
  \centering
  \caption{$k$-NN sweep on ZeMA (3-seed mean overall AUROC, with weak/severe).
  No-stop-gradient peaks at $k=20$, stop-gradient at $k=50$. Both collapse at
  $k=200$ as the neighbourhood approaches the bank size.}
  \label{tab:zema_ksweep}
  \begin{tabular}{lcccccc}
    \toprule
    & \multicolumn{3}{c}{Stop-gradient} & \multicolumn{3}{c}{No stop-gradient} \\
    \cmidrule(lr){2-4}\cmidrule(lr){5-7}
    $k$ & Overall & Weak & Severe & Overall & Weak & Severe \\
    \midrule
    1   & 0.8147 & 0.7371 & 0.8923 & 0.9709 & 0.9430 & 0.9988 \\
    5   & 0.8257 & 0.7515 & 0.9000 & 0.9717 & 0.9442 & 0.9993 \\
    10  & 0.8305 & 0.7584 & 0.9026 & 0.9723 & 0.9448 & 0.9998 \\
    20  & 0.9090 & 0.8447 & 0.9733 & \textbf{0.9843} & 0.9688 & 0.9998 \\
    50  & \textbf{0.9288} & 0.8710 & 0.9866 & 0.9561 & 0.9210 & 0.9912 \\
    100 & 0.9115 & 0.8449 & 0.9782 & 0.7967 & 0.6929 & 0.9005 \\
    200 & 0.3834 & 0.3378 & 0.4289 & 0.3857 & 0.3486 & 0.4229 \\
    \bottomrule
  \end{tabular}
\end{table}

\begin{table}[H]
  \centering
  \caption{CMR-Mamba on SWaT (3-seed mean $\pm$ std, window-level AUROC). The
  mechanism residual is the best CMR score{, and} $k$ is essentially flat.}
  \label{tab:swat_cmr}
  \begin{tabular}{lcc}
    \toprule
    Score & StopGrad & NoStopGrad \\
    \midrule
    $k$-NN, $k{=}1$   & 0.8620 & 0.8166 \\
    $k$-NN, $k{=}5$   & \textbf{0.8629} & 0.8172 \\
    $k$-NN, $k{=}20$  & 0.8615 & 0.8179 \\
    $k$-NN, $k{=}100$ & 0.8524 & \textbf{0.8186} \\
    $k$-NN, $k{=}200$ & 0.8461 & 0.8184 \\
    \midrule
    \textbf{mechanism residual} & $\mathbf{0.8679 \pm 0.0048}$ & $0.8557 \pm 0.0094$ \\
    \bottomrule
  \end{tabular}
\end{table}

\begin{table}[H]
  \centering
  \caption{Cause-reconstruction weight ($\alpha_{x}$) sweep on ZeMA. The optimum is $\alpha_{x}=0.00$.}
  \label{tab:zema_beta}
  \begin{tabular}{lccc}
    \toprule
    $\alpha_{x}$ & Overall & Weak & Severe \\
    \midrule
    \textbf{0.00} & \textbf{0.9977} & 0.9954 & 1.0000 \\
    0.25 & 0.9536 & 0.9071 & 1.0000 \\
    0.50 & 0.9217 & 0.8468 & 0.9965 \\
    0.75 & 0.9178 & 0.8417 & 0.9939 \\
    1.00 & 0.8262 & 0.7479 & 0.9044 \\
    \bottomrule
  \end{tabular}
\end{table}

\begin{table}[H] 
  \centering
  \caption{{Cause-reconstruction weight ($\alpha_{x}$)
  sweep on SWaT (single split, StopGrad). Both scores are flat across the range{, and}
  $\alpha_{x}=0$ is adopted (no actuator decoders needed).}}
  \label{tab:swat_beta}
  \begin{tabular}{lcc}
    \toprule
    {$\alpha_{x}$} & {$k$-NN AUROC} & {Mech-resid AUROC} \\
    \midrule
    {\textbf{0.00}} & {0.8668} & {0.8610} \\
    {0.25} & {0.8682} & {0.8612} \\
    {0.50} & {0.8693} & {0.8644} \\
    {0.75} & {0.8688} & {0.8646} \\
    {1.00} & {0.8697} & {0.8661} \\
    \bottomrule
  \end{tabular}
\end{table}
\section*{Declaration on the Use of Generative AI}
All research reported in this paper, including its conception, methodology, implementation, experiments, analysis, and the writing of the manuscript, was carried out by the authors. Generative AI tools, including GitHub and Copilot, were used only as assistive aids,i.e., to help resolve \LaTeX{} errors, to check syntax and formatting, to perform grammar and spelling checks, and to make individual paragraphs more concise. They were not used to generate research ideas, results, or claims. The authors reviewed all content and take full responsibility for it.
\bibliographystyle{ACM-Reference-Format}
\bibliography{refV}
% \balance

\end{document}